\documentclass[letterpaper]{article} 
\usepackage[preprint]{aaai2027}  
\usepackage[hyphens]{url}  
\usepackage{graphicx} 
\usepackage{natbib}  
\usepackage{caption} 
\usepackage{algorithm}
\usepackage{algorithmic}

\usepackage{newfloat}
\usepackage{listings}
\DeclareCaptionStyle{ruled}{labelfont=normalfont,labelsep=colon,strut=off} 
\floatstyle{ruled}
\newfloat{listing}{tb}{lst}{}
\floatname{listing}{Listing}

\usepackage{booktabs}

\title{AAAI Press Anonymous Submission\\Instructions for Authors Using \LaTeX{}}
\author{
    Written by AAAI Press Staff\textsuperscript{\rm 1}\thanks{With help from the AAAI Publications Committee.}\\
    AAAI Style Contributions by Peter Patel Schneider,
    Sunil Issar,\\
    J. Scott Penberthy,
    George Ferguson,
    Hans Guesgen,
    Francisco Cruz\equalcontrib\corresponding,
    Marc Pujol-Gonzalez\equalcontrib\corresponding
}
\affiliations{
    \textsuperscript{\rm 1}Association for the Advancement of Artificial Intelligence\\

    1101 Pennsylvania Ave, NW Suite 300\\
    Washington, DC 20004 USA\\
    proceedings-questions@aaai.org
}

\usepackage{tcolorbox}
\tcbuselibrary{skins,breakable}
\usepackage{tikz}
\usetikzlibrary{positioning}
\usepackage{algorithm}
\usepackage{algorithmic}
\usepackage{booktabs}
\usepackage{multirow}
\usepackage{xcolor}
\definecolor{best}{HTML}{1A73E8} 
\usepackage{cuted}

\usepackage{xcolor}      
\usepackage{graphicx}    
\usepackage{tabularx}    
\usepackage{booktabs}    
\usepackage{adjustbox}

\definecolor{okgreen}{RGB}{34,139,34}
\definecolor{failred}{RGB}{198,65,58}
\definecolor{warnorange}{RGB}{200,110,20}

\newcommand{\safeimg}[2]{\IfFileExists{#1}{\includegraphics[width=#2]{#1}}{\fbox{\parbox[c][2.6cm][c]{#2}{\centering\scriptsize image not found:\\ \texttt{#1}}}}}

\usepackage{newfloat}
\usepackage{listings}
\DeclareCaptionStyle{ruled}{labelfont=normalfont,labelsep=colon,strut=off} 
\floatstyle{ruled}
\newfloat{listing}{tb}{lst}{}
\floatname{listing}{Listing}
\usepackage{tcolorbox}
\tcbuselibrary{skins}
\definecolor{percC}{HTML}{185FA5}
\definecolor{groundC}{HTML}{0F6E56}
\definecolor{reasC}{HTML}{534AB7}
\definecolor{phC}{HTML}{BA7517}

\newtcolorbox{stagebox}[2]{%
  colframe=#1, colbacktitle=#1, colback=#1!6, coltitle=white,
  fonttitle=\bfseries, title=#2, boxrule=0.6pt, arc=3pt,
  left=5pt, right=5pt, top=3pt, bottom=3pt, before skip=4pt, after skip=4pt}

\usepackage{booktabs}      
\usepackage{xcolor}
\definecolor{impup}{RGB}{0,128,0}     
\definecolor{impdn}{RGB}{200,0,0}     

\newcommand{\gain}[2]{#1\textsubscript{\textcolor{impup}{$\uparrow$#2}}}
\newcommand{\drop}[2]{#1\textsubscript{\textcolor{impdn}{$\downarrow$#2}}}
\newcommand{\gainb}[2]{\textbf{#1}\textsubscript{\textcolor{impup}{$\uparrow$#2}}}

\usepackage{amsmath}
\usepackage{amssymb}
\usepackage{booktabs}
\usepackage{tabularx}
\usepackage{array}
\usepackage[table]{xcolor}

\definecolor{flipfg}{HTML}{0F6E56}\definecolor{flipbg}{HTML}{E1F5EE}
\definecolor{missfg}{HTML}{C0392B}\definecolor{missbg}{HTML}{FCEBEB}
\definecolor{okfg}{HTML}{5F5E5A}\definecolor{okbg}{HTML}{F1EFE8}
\definecolor{fmtfg}{HTML}{BA7517}

\usepackage{graphicx}
\usepackage[table]{xcolor}     
\usepackage{booktabs}
\usepackage{tabularx}
\usepackage{tcolorbox}
\definecolor{phC}{HTML}{7A4CC2}
\definecolor{okC}{HTML}{1B7A3D}
\definecolor{missC}{HTML}{B00020}
\definecolor{fmtC}{HTML}{B36B00}
\definecolor{cprow}{HTML}{DDE5EF}   

\title{ChartProbe: A Diagnostic Study on Visual Reasoning through \\ Perception, Grounding, and Simple Reasoning}

\author{
    Written by AAAI Press Staff\textsuperscript{\rm 1}\thanks{With help from the AAAI Publications Committee.}\\
    AAAI Style Contributions by Pater Patel Schneider,
    Sunil Issar,\\
    J. Scott Penberthy,
    George Ferguson,
    Hans Guesgen,
    Francisco Cruz\equalcontrib,
    Marc Pujol-Gonzalez\equalcontrib
}
\affiliations{
    \textsuperscript{\rm 1}Association for the Advancement of Artificial Intelligence\\

    1101 Pennsylvania Ave, NW Suite 300\\
    Washington, DC 20004 USA\\
    proceedings-questions@aaai.org
}

\author {
    Mahsa Khoshnoodi,
    Sarah Adel Bargal
}
\affiliations {
    Department of Computer Science, Georgetown University\\
}

\usepackage{bibentry}

\begin{document}

\maketitle

\begin{abstract}
Vision-language models (VLMs) remain unreliable on chart questions that require
reasoning over visual quantities, and this weakness is usually attributed to a
reasoning deficit and addressed with more reasoning supervision. We ask whether
the difficulty lies in reasoning itself, or in the simpler skills that
reasoning operates on: reading the plotted elements (\emph{perception}),
locating them and binding them to their labels (\emph{grounding}), and
performing single-step computations such as ranking, totals, and differences
(\emph{simple reasoning}). 
We introduce \textbf{ChartProbe}, a diagnostic framework whose probes are generated directly from the code that renders each chart, so every gold answer is exact by construction, needs no human annotation, and attributes each failure to a single skill. ChartProbe enables an intervention prior work does not attempt: instead of synthesizing complex-reasoning data, we withhold complex questions and reasoning traces entirely, fine-tune on one simple skill at a time, and measure transfer to held-out complex-reasoning questions. Across three open-weight VLMs,
supervising the simpler skills alone produces large gains on complex-reasoning
questions the model never trained on: where these skills are weak and the model
can be taught to read the image, training them recovers much of complex
reasoning at no reasoning-data cost. The gains hold across three
out-of-distribution settings: an unseen chart type (pie charts), a
human-written benchmark disjoint from our images and templates (ChartQA), and a
non-chart visual domain (CLEVR). Complex visual reasoning can therefore improve
without complex-reasoning supervision.
\end{abstract}

\section{Introduction}
\label{sec:intro}


When a vision-language model (VLM) answers a chart question incorrectly, the default explanation is often that it lacks reasoning ability. This paper examines whether that explanation is actually correct. VLMs are increasingly expected to interpret charts and answer questions about trends, magnitudes, and relationships that are available only through visual evidence \citep{masry2022chartqa}. Yet their performance remains unreliable on tasks that require combining multiple values, comparing segments, ranking categories, or computing derived quantities such as differences from an average. The prevailing response has been to improve reasoning directly through chain-of-thought prompting, supervised fine-tuning on curated multi-step reasoning traces, reinforcement learning with reasoning-based rewards, or simply scaling the amount of complex-reasoning supervision.


Answering a chart question, however, is not a monolithic reasoning task. Instead, it can be decomposed into three fundamental skills: extracting visual attributes such as plotted elements, colors, and magnitudes (Perception); associating those elements with the corresponding labels and legend entries they represent (Grounding); and performing simple operations on the recovered values (Simple Reasoning). Each skill is relatively straightforward in isolation, unlike the multi-step deliberation targeted by chain-of-thought and reasoning-focused training. Consequently, an incorrect answer may arise from a failure in any one of these components, while aggregate accuracy or a black-box benchmark score provides little insight into the underlying cause. Treating every error as a reasoning failure therefore risks misdiagnosis, directing effort toward improving capabilities that are already sufficient while leaving the true bottleneck unaddressed.

\begin{figure*}[!t]
\centering
\includegraphics[width=\textwidth]{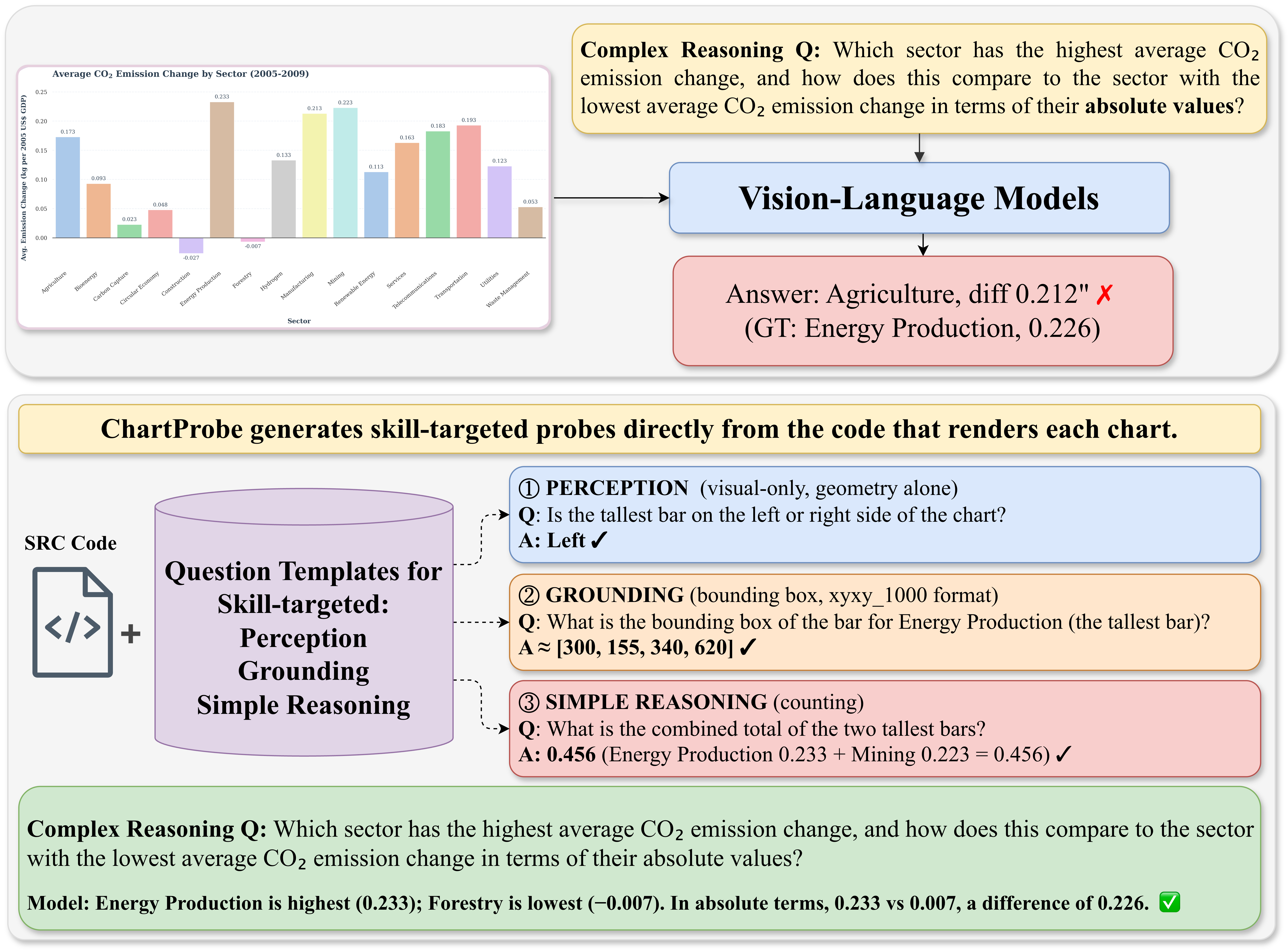}
\caption{\textbf{Top:} a vision-language model answers a complex chart
question incorrectly, naming the wrong sectors and difference. \textbf{Bottom:}
ChartProbe generates three skill-targeted probes directly from the chart's
source code and a bank of question templates: a \emph{perception} probe
answerable from the image alone, a \emph{grounding} probe that binds a named
sector to its bar, and a \emph{simple-reasoning} probe that computes over the
read values. Each probe has an exact answer known from the rendering code. The
model answers all three correctly, and answers the complex question correctly
once these skills are supplied, even though the complex question never appears
in training.}
\label{fig:pipeline}
\end{figure*}

We introduce \textbf{ChartProbe}, a diagnostic probe set that makes this
decomposition directly measurable (Figure~\ref{fig:pipeline}). For each chart, ChartProbe
instantiates a fixed bank of skill-targeted probes derived from question templates
spanning the three atomic skills: perception, grounding, and simple reasoning. Each probe
is generated deterministically from the code and data that render the chart, so
its ground truth is exact by construction. As a result, ChartProbe requires neither human annotation nor model-generated labels, both of which introduce errors of their own. 


Using lightweight atomic-skill fine-tuning, we supervise individual skills—or selected combinations of skills—while withholding all complex-reasoning data during training. We then evaluate the resulting models on the held-out set of complex reasoning questions associated with a separate collection of charts. This setup allows us to directly measure how improvements in specific foundational skills transfer to complex reasoning performance without any explicit supervision on the target task.


The logic is inherently interventional: if performance on complex reasoning tasks is primarily constrained by multi-step reasoning, then only reasoning-specific supervision should improve it. Conversely, if performance is limited by the underlying skills on which reasoning depends, then improving perception and/or grounding should also lead to gains. Across three open-weight vision-language models: InternVL \cite{wang2025internvl3}, Qwen \cite{team2026qwen3}, and LLaVA \cite{liu2023improvedllava}, we find strong evidence for the latter. 

We find that complex reasoning performance improves substantially even when training is restricted to simpler skills and never directly targets complex reasoning itself. These gains generalize across three out-of-distribution settings: an unseen chart type (pie charts, excluded from training), a human-authored benchmark with disjoint images and questions (ChartQA \cite{masry2022chartqa}), and a non-chart visual domain (CLEVR scenes \cite{johnson2017clevr}), demonstrating that the effect is not specific to a particular dataset or task distribution.

\noindent We summarize our contributions as follows:
\begin{itemize}
\item We introduce \textbf{ChartProbe}, a diagnostic that generates skill-targeted probes directly from the code that renders each chart, decomposing visual question answering into three skills: perception, grounding, and simple reasoning. Because ground truth is programmatic, every gold answer is exact by construction and needs no human annotation. ChartProbe spans bar and pie charts and extends to CLEVR scenes, giving one probe generator across visual domains.
\item We show that fine-tuning on these simple skills alone, with no complex questions or reasoning traces in training, transfers to held-out complex reasoning across multiple open-weight vision-language models. Simple-skill supervision alone recovers much of the complex-reasoning accuracy that models otherwise reach only with direct complex-reasoning training.
\item Models trained only on this automatically generated simple-skill data generalize to three out-of-distribution settings: an unseen chart type (pie charts), a human-authored benchmark disjoint from our images and templates (ChartQA), and a non-chart visual domain (CLEVR). These probe distinct forms of distribution shift, showing the gains are not tied to a single axis of generalization.
\end{itemize}


\section{Related Works}
\label{sec:related}


\paragraph{Reasoning supervision in vision-language models.}
Reasoning is widely treated as the capability separating strong from weak
multimodal models. Chain-of-thought prompting and its multimodal extensions
elicit intermediate steps on compositional tasks
\cite{wei2022chain,zhang2023multimodal}, and reinforcement-learned models port
the DeepSeek-R1 recipe into vision-language training to elicit long deliberate
traces \cite{chen2025g1,yang2025r1,huang2025vision}. These methods operate on
whatever the model has already encoded, and typically do not verify that the
visual evidence entering the reasoning step was read correctly. Our results
suggest that this assumption does not always hold, and that supervision
directed at reasoning can leave the limiting skills untouched.

\paragraph{Perception-centric interventions.}
A second line intervenes on perception rather than reasoning. Vision-language
models struggle with visual distinctions that are trivial for humans, and the
difficulty traces to image-encoder representations that collapse visibly
distinct inputs \cite{rahmanzadehgervi2024vision,tong2024eyes}. Two families of
methods act on this. One routes around the encoder, extracting a structured
representation of the image with a dedicated module and delegating computation
to a language model \cite{liu2023matcha,liu2023deplot}. The other strengthens
perception in place, injecting it as a reward \cite{xiao2025perception}, as
grounding supervision \cite{yu2026perception}, or through tool-based reasoning
over structured inputs \cite{wu2025vtool}. Both raise accuracy, but neither shows where the original model went wrong. One replaces the suspect component; the other changes the model and its accuracy at the same time. ChartProbe instead
measures each skill separately in the unmodified model, so the intervention and
the measurement stay distinct.

\paragraph{Skill-targeted probing and transfer.}
Behavioral decompositions of visual questions pose skill-targeted
sub-questions with verifiable answers, in contrast to probing that trains
classifiers on internal activations
\cite{alain2016understanding,tenney2019bert}. Such decompositions draw their
intermediate units from human-annotated sub-questions
\cite{selvaraju2020squinting}, symbolic parsing \cite{yi2018neural}, or model
generation \cite{jin2026seeing,li2025self,lee2026visdot}, and chart benchmarks
have begun to separate descriptive from reasoning questions
\cite{wang2024charxiv}. A parallel line uses such decompositions to synthesize
training data: COGS \cite{gu2025composition} prompts an MLLM to factor seed
questions into perception and reasoning components, recomposes them with new
images to generate question-answer pairs, whose intermediate answers come from
the same model, and supervises those subquestions as process rewards while
training on complex questions. Across this work the decomposition serves the
target task: intermediate units are used to score models or to build training
data, and complex supervision is retained throughout. We instead treat the
decomposition as an intervention, supervising one skill in isolation while
withholding complex questions entirely, which isolates what simple-skill
supervision contributes on its own.

\paragraph{Charts as a controlled substrate.}
Chart understanding has dedicated benchmarks
\cite{masry2022chartqa,methani2020plotqa,masry2025chartqapro} and specialized
architectures \cite{masry2025chartgemma,han2023chartllama}, but our goal is not
to advance chart processing. We use charts because they supply what skill
isolation requires and natural images do not: generating code that fixes the
perceptual input, the label-to-mark mapping, and the expected computation. This
control has long motivated synthetic chart QA
\cite{kahou2017figureqa,kafle2018dvqa}, where known plotting data yields exact
answers; we use it not to build a benchmark but to isolate and supervise one
skill at a time. To ensure the effect is not confined to synthetic charts, we evaluate on human-written ChartQA questions \cite{masry2022chartqa}, authored independently of our templates.
\section{ChartProbe}
\label{sec:method}

\label{subsec:selection}
Each probe in ChartProbe has an answer computed from the chart's generating
code rather than inferred from the image, which keeps model error separate from
any error in the ground truth. Because our probes depend on source-level ground truth, we use ChartNet~\cite{kondic2026chartnet}, a dataset that pairs every rendered chart with its generating Python code and CSV data files. We train only on
single-series bar charts, excluding multi-series charts to avoid structural
confounds, and apply the same decomposition to pie charts at evaluation time as
a held-out chart type. The training pool is 1{,}000 bar charts stratified over
three complexity tiers by bar count: simple (3--5), medium (6--10), and complex
(11--24).
\paragraph{Held-out sets.} The probe test sets contain 500 bar charts,
500 pie charts, and 500 CLEVR scenes, all disjoint from the training
pool; the CLEVR probes are generated programmatically from the original
scene graphs~\cite{johnson2017clevr}. For complex reasoning we use two
question sources. From ChartNet's reasoning split~\cite{kondic2026chartnet}
we take 500 bar and 500 pie questions (one per chart) over charts held out
from the probe pool, and CLEVR contributes 500 questions from its original
test set. To measure generalization to real-world charts, ChartQA~\cite{masry2022chartqa}
adds 500 bar and 500 pie questions from an independent, human-authored
dataset. This gives 500 questions per chart type per source.

\subsection{Skill-Targeted Probes}
\label{subsec:questions}
A probe is a programmatic map $P(S) \rightarrow (Q, A)$ from a chart's source
specification $S$ to a question $Q$ and an answer $A = f(S)$ computed from source
primitives. For charts, $S$ comprises the generating Python code and its CSV data
file; for CLEVR, the scene-graph JSON. Each template pairs a natural-language question schema with an answer function over $S$: instantiating it on an instance
fills the schema with that instance's primitives (for charts, its column headers
and values; for CLEVR, its objects and attributes) and evaluates the function, so the ground truth is exact and requires no annotation. The templates are
hand-authored, and each targets a single skill.
Perception templates ask for judgments readable from the rendered image alone, with no value read from any axis: for bar charts, bar counts, height comparisons, extremal position, left-to-right trend, and color distinctness. Grounding templates ask for label-to-element and element-to-label binding, for the value of a named element, and for bounding-box prediction. Simple-reasoning templates ask for a single operation over the values: a ranking, sum, mean, difference, or percentage. The complete template bank is given for bar charts in Supplementary material; the
pie bank follows the same design, adapting each template to segments, and CLEVR uses parallel templates over scene-graph primitives.
\paragraph{Targeting criterion.} Each probe is built to test one atomic skill, so
that a wrong answer is attributable to that skill rather than left ambiguous. This
is a property of what a probe \emph{measures}. \emph{Perception} probes are genuinely self-contained: they
require only relative magnitude judgments readable from the image, with no numeric value extracted and no category name referenced.
\emph{Grounding} probes test the binding of a language label to a visual element, in both directions, together with normalized bounding-box prediction verified
against coordinates from the plotting backend. \emph{Simple-reasoning} probes test a single computation over chart values, such as ranking bars, totalling them, or
differencing two named bars. The criterion is which skill a probe targets, not how hard it is: a
perception probe may involve comparison against an aggregate magnitude, provided
that comparison is available visually without reading a value off the axis.


\paragraph{Transfer domains.} The evaluation spans four domains chosen to separate a transferable skill from a dataset-specific shortcut. Bar charts are in-distribution, sharing chart type, source, and rendering pipeline with training, and measure fit to the supervised setting. Pie charts are an unseen chart type but share the source: they inherit ChartNet's generation process while presenting geometry the model never saw supervised, and measure transfer when only the visual form changes. ChartQA is an unseen question source, comprising real-world charts from a separate dataset with independent
rendering and annotation, so a gain there cannot be credited to ChartNet-specific
regularities of layout, style, or question phrasing. CLEVR is an unseen visual
domain, containing no bars, axes, or chart structure, and tests whether a supervised atomic skill is a chart-specific shortcut or a transferable visual competence. Comparing
results across these four settings separates genuine skill transfer from
distributional overlap: an effect confined to the in-distribution setting reflects
a memorized regularity, whereas one that holds on ChartQA and CLEVR is evidence that
the supervised skill generalizes to complex reasoning beyond the training domain.
\paragraph{Atomic skill evaluation.}
\label{subsec:eval}
Each skill probe has a closed-form ground truth, so we score it with a matcher
suited to its answer type rather than a single global metric. Binary, count, and
label questions are scored by exact match after lowercasing and whitespace
stripping, since their answers occupy a small, unambiguous space. Value questions
use a $\pm5\%$ relative tolerance following the ChartQA
protocol~\cite{masry2022chartqa}, crediting a read that is numerically correct up
to minor rounding or axis-interpolation error. List questions, whose answers are
unordered sets, use recall against the ground truth, and bounding-box questions use
intersection-over-union at a $0.5$ threshold, the standard localization criterion.
Because every target is computed from the chart source, these matchers compare
against exact ground truth and introduce no annotation noise. All models are decoded greedily with a fixed generation budget.

\paragraph{Complex-reasoning evaluation.}
Complex-reasoning answers are free-form generations, where exact match penalizes
responses that are semantically correct but lexically divergent, and a hard
threshold discards the graded information in near-misses. This is well documented
for open-ended VLM evaluation, where instruction-tuned models emit verbose or
reformatted answers that a strict match scores as wrong. We therefore report the mean RapidFuzz \texttt{ratio}~\cite{rapidfuzz}, a normalized
edit-distance similarity in $[0,1]$ between the extracted answer and the ground truth, following prior fuzzy-matching practice~\cite{kondic2026chartnet}. This is further motivated by the error structure we observe:
many near-misses are correct in content but diverge in surface form, emitting a sentence or an inline reasoning trace where a single token was expected, so a strict
match would score formatting rather than correctness. 
\section{Experiments}
\label{sec:experiments}


We design experiments to test complex reasoning both in- and out-of-domain. In-domain complex reasoning is evaluated on ChartNet bar-chart samples; out-of-domain complex reasoning is evaluated on (1) ChartNet pie charts, (2) ChartQA, a separate chart dataset, and (3) CLEVR scenes. Because supervision is confined to bar charts, the pie-chart and CLEVR results are out of distribution, and ChartQA is additionally out of source. Fine-tuning numbers are means over three seeds. All configurations use an identical fine-tuning format, so differences between them isolate the effect of which skills are supervised, independent of answer form.

\paragraph{Setup.} We fine-tune three open-weight VLMs, InternVL-3.5-8B,
Qwen3.5-9B, and LLaVA-1.6-Mistral-7B, with LoRA adapters (rank \(r=32\),
\(\alpha=64\)) applied to the attention and MLP projections, leaving the vision
encoder frozen. Each model is trained under five skill-set configurations: three
single-skill, Perception (\emph{P}), Grounding (\emph{G}), and Simple Reasoning
(\emph{SR}), and two combined (\emph{P+G}, \emph{P+G+SR}), none of which contains a
complex-reasoning question. Every configuration is trained for \(N=1\) epoch with a
learning rate of \(1\times10^{-4}\) (InternVL, LLaVA) or \(2\times10^{-4}\) (Qwen)
and an effective batch size of \(8\) (InternVL, LLaVA) or \(16\) (Qwen), on a 90/10
train/validation split, with hyperparameters held fixed across the five
configurations for each model so that only the supervised skill set varies. Every
run is repeated over three seeds (42/43/44) and we report the mean; the base model
is evaluated zero-shot over three inference repetitions. All experiments run on
4\(\times\) L40S GPUs, with greedy decoding and a fixed generation budget at
evaluation..

\paragraph{Results.} Table~\ref{tab:chartprobe_finetune} shows no single failure profile. On bar charts InternVL is weakest in perception (36.8, against 52.1 grounding, and 53.5 simple reasoning), whereas Qwen is weakest in perception and simple reasoning alike (15.2, 12.6). On Qwen's pie charts, the simple-reasoning probe starts at 11.8, and perception-plus-grounding supervision, with no reasoning data, raises it to 64.9, matching the 63.5 that direct reasoning supervision reaches. CLEVR shows the same pattern in a non-chart domain, where every model's simple-reasoning probe is weakest at baseline (13.6 for InternVL) and recovers under supervision that never touches reasoning. The effect is bounded, not universal: where simple reasoning is already strong, as for InternVL on charts (53.5, 54.9), there is little to recover and perception-only supervision can even lower it slightly. LLaVA is weak across all three skills on both chart types and is the boundary case we return to below. Because supervising multiple skills increases the number of training examples, the combined configurations could gain from more data rather than broader skill coverage. Repeating the comparison with total training examples held fixed across configurations yields similar trends, i.e. the transfer effect is not explained by volume alone. Results for this experiment can be found in the supplementary material.
\begin{table*}[t]
\centering
\caption{Skill-wise diagnostic accuracy (\%) on ChartProbe-\textbf{500} ($n{=}4500$ questions). Rows are the base model and five LoRA configurations: single-skill \emph{P}, \emph{G}, \emph{SR} and combined \emph{P+G}, \emph{P+G+SR}. Fine-tuned rows are mean over 3 seeds; the base model is mean over 3 evaluation reps. Columns \textbf{P}/\textbf{G}/\textbf{SR} are the perception, grounding, and simple-reasoning probes. Subscripts give absolute change over the base model (\textcolor{impup}{$\uparrow$} gain, \textcolor{impdn}{$\downarrow$} drop). Best per column within each dataset block in \textbf{bold}.}
\label{tab:chartprobe_finetune}
\setlength{\tabcolsep}{8pt}\small
\begin{tabular}{l ccc ccc ccc}
\toprule
& \multicolumn{3}{c}{InternVL-3.5-8B} & \multicolumn{3}{c}{Qwen3.5-9B} & \multicolumn{3}{c}{LLaVA-1.6 Mistral-7B} \\
\cmidrule(lr){2-4} \cmidrule(lr){5-7} \cmidrule(lr){8-10}
Train & P & G & SR & P & G & SR & P & G & SR \\
\midrule
\rowcolor{cprow}\multicolumn{10}{l}{\textit{bar charts}} \\
Base    & 36.8 & 52.1 & 53.5 & 15.2 & 25.6 & 12.6 & 0.3 & 26.6 & 6.5 \\
\emph{P}     & \gain{85.9}{49.1} & \gain{55.6}{3.5} & \drop{49.0}{4.5} & \gain{57.4}{42.2} & \gain{39.5}{13.9} & \gain{50.2}{37.6} & \gain{71.6}{71.3} & \gain{31.4}{4.8} & \drop{5.8}{0.7} \\
\emph{G}     & \gain{42.3}{5.5} & \gain{\textbf{71.6}}{19.5} & \drop{35.6}{17.9} & \gain{32.6}{17.4} & \gain{38.6}{13.0} & \gain{29.5}{16.9} & \gain{29.9}{29.6} & \gain{64.4}{37.8} & \drop{5.8}{0.7} \\
\emph{SR}     & \gain{46.2}{9.4} & \gain{53.8}{1.7} & \gain{68.4}{14.9} & \gain{43.8}{28.6} & \gain{45.7}{20.1} & \gain{63.5}{50.9} & \gain{31.5}{31.2} & \gain{39.8}{13.2} & \gain{44.7}{38.2} \\
\emph{P+G}   & \gain{85.2}{48.4} & \gain{66.7}{14.6} & \drop{43.4}{10.1} & \gain{\textbf{78.6}}{63.4} & \gain{\textbf{55.5}}{29.9} & \gain{55.0}{42.4} & \gain{72.3}{72.0} & \gain{64.2}{37.6} & \gain{7.7}{1.2} \\
\emph{P+G+SR} & \gain{\textbf{86.6}}{49.8} & \gain{68.4}{16.3} & \gain{\textbf{69.0}}{15.5} & \gain{66.3}{51.1} & \gain{46.0}{20.4} & \gain{\textbf{64.9}}{52.3} & \gain{\textbf{73.6}}{73.3} & \gain{\textbf{65.5}}{38.9} & \gain{\textbf{45.3}}{38.8} \\
\midrule
\rowcolor{cprow}\multicolumn{10}{l}{\textit{pie charts}} \\
Base    & 42.6 & 61.5 & 54.9 & 12.7 & 31.2 & 11.8 & 0.2 & 22.3 & 11.0 \\
\emph{P}     & \gain{73.2}{30.6} & \drop{61.2}{0.3} & \gain{57.6}{2.7} & \gain{49.3}{36.6} & \gain{49.0}{17.8} & \gain{56.4}{44.6} & \gain{59.6}{59.4} & \gain{29.8}{7.5} & \gain{11.2}{0.2} \\
\emph{G}     & \drop{39.7}{2.9} & \drop{58.8}{2.7} & \drop{49.6}{5.3} & \gain{36.5}{23.8} & \gain{45.5}{14.3} & \gain{27.8}{16.0} & \gain{32.4}{32.2} & \gain{\textbf{35.3}}{13.0} & \gain{12.3}{1.3} \\
\emph{SR}     & \gain{43.5}{0.9} & \drop{60.7}{0.8} & \gain{61.0}{6.1} & \gain{47.0}{34.3} & \gain{57.3}{26.1} & \gain{63.5}{51.7} & \gain{28.6}{28.4} & \gain{29.1}{6.8} & \gain{\textbf{32.9}}{21.9} \\
\emph{P+G}   & \gain{\textbf{74.7}}{32.1} & \drop{61.0}{0.5} & \gain{57.2}{2.3} & \gain{\textbf{68.5}}{55.8} & \gain{\textbf{64.0}}{32.8} & \gain{64.9}{53.1} & \gain{\textbf{63.8}}{63.6} & \gain{34.5}{12.2} & \drop{3.0}{8.0} \\
\emph{P+G+SR} & \gain{70.5}{27.9} & \gain{\textbf{61.7}}{0.2} & \gain{\textbf{63.1}}{8.2} & \gain{51.3}{38.6} & \gain{46.3}{15.1} & \gain{\textbf{68.3}}{56.5} & \gain{56.5}{56.3} & \gain{32.9}{10.6} & \gain{25.0}{14.0} \\
\midrule
\rowcolor{cprow}\multicolumn{10}{l}{\textit{CLEVR}} \\
Base    & 86.1 & 45.9 & 13.6 & 43.8 & 26.1 & 21.6 & 3.3 & 3.9 & 1.1 \\
\emph{P}     & \gain{90.6}{4.5} & \gain{54.5}{8.6} & \gain{80.1}{66.5} & \gain{81.9}{38.1} & \gain{52.4}{26.3} & \gain{79.5}{57.9} & \gain{23.4}{20.1} & \gain{34.9}{31.0} & \gain{38.2}{37.1} \\
\emph{G}     & \gain{89.7}{3.6} & \gain{53.5}{7.6} & \gain{85.5}{71.9} & \gain{52.7}{8.9} & \gain{37.1}{11.0} & \gain{55.7}{34.1} & \gain{\textbf{33.3}}{30.0} & \gain{40.3}{36.4} & \gain{52.4}{51.3} \\
\emph{SR}     & \gain{95.0}{8.9} & \gain{62.2}{16.3} & \gain{\textbf{91.2}}{77.6} & \gain{71.4}{27.6} & \gain{52.7}{26.6} & \gain{63.2}{41.6} & \gain{31.5}{28.2} & \gain{45.0}{41.1} & \gain{52.1}{51.0} \\
\emph{P+G}   & \gain{89.4}{3.3} & \gain{58.6}{12.7} & \gain{88.2}{74.6} & \gain{\textbf{85.0}}{41.2} & \gain{\textbf{63.9}}{37.8} & \gain{\textbf{79.7}}{58.1} & \gain{29.2}{25.9} & \gain{\textbf{46.2}}{42.3} & \gain{41.7}{40.6} \\
\emph{P+G+SR} & \gain{\textbf{95.8}}{9.7} & \gain{\textbf{63.6}}{17.7} & \gain{90.6}{77.0} & \gain{85.0}{41.2} & \gain{55.1}{29.0} & \gain{66.4}{44.8} & \gain{27.7}{24.4} & \gain{45.3}{41.4} & \gain{\textbf{53.6}}{52.5} \\
\bottomrule
\end{tabular}
\end{table*}

\paragraph{Skills are separately trainable, and they interact.}
Each skill's own supervision produces its largest in-distribution gain on that skill, confirming the skills are separable targets: for InternVL on bar charts, perception supervision raises perception by 49.1 points, grounding supervision raises grounding by 19.5, and simple-reasoning supervision raises simple reasoning by 14.9. They also interact. On CLEVR, perception-only supervision lifts InternVL's simple-reasoning probe from 13.6 to 80.1 and Qwen's from 21.6 to 79.5; for Qwen this exceeds the 63.2 that simple-reasoning supervision reaches on the same probe, so on this probe supervising perception alone helps reasoning more than supervising reasoning does. Perception also contributes beyond grounding: on Qwen's CR-CLEVR, adding perception to grounding supervision lifts complex reasoning 24.9 points over grounding alone, though both train on the same grounding data.

The interaction runs both ways, and we report it plainly. In distribution, single-skill supervision can reduce another skill: for InternVL on bar charts, perception-only supervision costs 4.5 points of simple reasoning and grounding-only costs 17.9. The combined \emph{P+G} and \emph{P+G+SR} configurations largely avoid this trade-off, which is why we treat multi-skill supervision as the default.

\paragraph{Complex reasoning (CR) improves under every form of simple-skill supervision.}
The decisive test is the held-out CR sets: compositional questions no configuration is trained on, in a format none of our templates produce. Across three models and three domains, nearly every form of simple-skill supervision improves complex reasoning over baseline (Table~\ref{tab:chartprobe_complex}). Two things make this hard to explain away as memorized transfer. First, the improvement is uniform across supervision type: complex reasoning rises under perception alone, grounding alone, simple reasoning alone, and their combinations, though none supplies a multi-step example. Second, the largest gains appear on pie and CLEVR, domains for which no training data exists at all, so what improves there cannot be recalled from training and must instead be a skill that generalizes. Complex reasoning therefore improves without any complex-reasoning supervision.

\begin{table*}[t]
\centering
\caption{Complex-reasoning accuracy (fuzzy \%) on the held-out \textbf{CR-500} sets (bar / pie / CLEVR, 500 questions each). Rows are the base model and five LoRA configurations: single-skill \emph{P}, \emph{G}, \emph{R} and combined \emph{P+G}, \emph{P+G+SR}; none is trained on any complex-reasoning question. Fine-tuned rows are mean over 3 seeds. Subscripts give absolute change over the base model. Best per column in \textbf{bold}.}
\label{tab:chartprobe_complex}
\setlength{\tabcolsep}{8pt}\small
\begin{tabular}{l ccc ccc ccc}
\toprule
& \multicolumn{3}{c}{InternVL-3.5-8B} & \multicolumn{3}{c}{Qwen3.5-9B} & \multicolumn{3}{c}{LLaVA-1.6 Mistral-7B} \\
\cmidrule(lr){2-4} \cmidrule(lr){5-7} \cmidrule(lr){8-10}
Train & Bar & Pie & CLEVR & Bar & Pie & CLEVR & Bar & Pie & CLEVR \\
\midrule
Base    & 24.0 & 26.0 & 48.0 & 8.6 & 9.0 & 20.2 & 17.3 & 18.5 & 7.0 \\
\emph{P}     & \gainb{36.7}{12.7} & \gainb{40.9}{14.9} & \gain{64.5}{16.5} & \gain{17.6}{9.0} & \gain{25.8}{16.8} & \gain{42.6}{22.4} & \gain{22.2}{4.9} & \gain{20.4}{1.9} & \gain{23.6}{16.6} \\
\emph{G}     & \gain{36.5}{12.5} & \gain{39.4}{13.4} & \gainb{67.1}{19.1} & \gain{13.3}{4.7} & \gain{18.1}{9.1} & \gain{32.3}{12.1} & \gain{19.4}{2.1} & \drop{17.4}{1.1} & \gainb{31.1}{24.1} \\
\emph{SR}     & \gain{36.6}{12.6} & \gain{39.5}{13.5} & \gain{64.5}{16.5} & \gain{23.2}{14.6} & \gain{28.8}{19.8} & \gain{42.3}{22.1} & \gain{25.9}{8.6} & \gain{20.4}{1.9} & \gain{30.0}{23.0} \\
\emph{P+G}   & \gain{28.5}{4.5} & \gain{34.6}{8.6} & \gain{62.4}{14.4} & \gainb{28.1}{19.5} & \gainb{36.6}{27.6} & \gainb{57.2}{37.0} & \gain{18.9}{1.6} & \drop{17.4}{1.1} & \gain{23.4}{16.4} \\
\emph{P+G+SR} & \gain{35.6}{11.6} & \gain{40.6}{14.6} & \gain{62.6}{14.6} & \gain{24.7}{16.1} & \gain{32.3}{23.3} & \gain{45.7}{25.5} & \gainb{26.2}{8.9} & \gainb{24.2}{5.7} & \gain{23.0}{16.0} \\
\bottomrule
\end{tabular}
\end{table*}

The effect we describe presupposes a model that can already read the chart; LLaVA shows what happens when one cannot. Its bar and pie perception baselines (0.3, 0.2) indicate a model that cannot deliver a usable read of the chart; for example, on bar charts it locates almost no bar correctly (perception 0.3). Its complex-reasoning gains on charts are correspondingly the smallest we observe, including the only two regressions in the table. CLEVR is the telling contrast: there LLaVA's perception is not floor-bound, and simple-skill supervision recovers $+24.1$, the same effect we see in the stronger models. The limit is therefore not LLaVA itself but whether its earlier skills can be trained at all: where a model cannot read the input, supervising simple skills has nothing to build on and complex reasoning does not move; where it can, the effect returns.

\paragraph{Inference on an independent benchmark.}
Thus far, our probes and CR sets are both built over ChartNet, so gains could reflect source-specific competence. We therefore evaluate all configurations on ChartQA~\citep{masry2022chartqa}, which shares no questions, templates, or charts with our training data, restricted to 500 bar and 500 pie charts and scored under the same fuzzy accuracy protocol, with no ChartQA data used in training. Every model improves on both chart types under every configuration (Table~\ref{tab:chartqa}). Gains are modest for the already-strong InternVL and large for the weaker baselines. Which skill helps most differs from the CR sets: here simple-reasoning
supervision is best in five of six columns, though perception supervision
still delivers $+32.0$ on Qwen bar and $+21.5$ on LLaVA bar.
Simple-reasoning supervision best matches ChartQA's short-answer format,
which likely inflates its edge, but perception and grounding supervision,
which share none of that format, also improve ChartQA on every model and
both chart types, so the transfer is not only stylistic.

\begin{table*}[t]
\centering
\caption{\textbf{ChartQA-500} off-the-shelf transfer accuracy (fuzzy \%; RapidFuzz). Rows are the base model and five LoRA configurations (single-skill \emph{P}, \emph{G}, \emph{R} and combined \emph{P+G}, \emph{P+G+SR}), evaluated on held-out ChartQA bar/pie 500 and never fine-tuned on ChartQA. Fine-tuned rows are mean over seeds 42/43/44; subscripts give absolute change over the base model (\textcolor{impup}{$\uparrow$} gain, \textcolor{impdn}{$\downarrow$} drop). Best per column in \textbf{bold}.}
\label{tab:chartqa}
\setlength{\tabcolsep}{8pt}\small
\begin{tabular}{l cc cc cc}
\toprule
& \multicolumn{2}{c}{InternVL-3.5-8B} & \multicolumn{2}{c}{Qwen3.5-9B} & \multicolumn{2}{c}{LLaVA-1.6 Mistral-7B} \\
\cmidrule(lr){2-3} \cmidrule(lr){4-5} \cmidrule(lr){6-7}
Train & Bar & Pie & Bar & Pie & Bar & Pie \\
\midrule
Base    & 76.9 & 74.2 & 39.7 & 48.2 & 25.0 & 28.4 \\
\emph{P}     & \gain{\textbf{83.7}}{6.8} & \gain{85.3}{11.1} & \gain{71.7}{32.0} & \gain{66.0}{17.8} & \gain{46.5}{21.5} & \gain{45.4}{17.0} \\
\emph{G}     & \gain{77.2}{0.3} & \gain{80.3}{6.1} & \gain{62.4}{22.7} & \gain{66.2}{18.0} & \gain{55.9}{30.9} & \gain{56.3}{27.9} \\
\emph{R}     & \gain{83.2}{6.3} & \gain{\textbf{87.0}}{12.8} & \gain{\textbf{78.8}}{39.1} & \gain{\textbf{79.3}}{31.1} & \gain{\textbf{58.2}}{33.2} & \gain{\textbf{58.9}}{30.5} \\
\emph{P+G}   & \gain{79.4}{2.5} & \gain{80.4}{6.2} & \gain{78.0}{38.3} & \gain{77.9}{29.7} & \gain{54.7}{29.7} & \gain{55.3}{26.9} \\
\emph{P+G+SR} & \gain{82.2}{5.3} & \gain{84.5}{10.3} & \gain{67.3}{27.6} & \gain{68.6}{20.4} & \gain{56.8}{31.8} & \gain{58.4}{30.0} \\
\bottomrule
\end{tabular}
\end{table*}

\paragraph{Chart specialization does not substitute for simple skills.}
We finally ask whether chart-specific pretraining closes the same gap. Table~\ref{tab:cr_ref} places two chart-oriented reference models against our skill-tuned models on the same CR sets. ChartGemma~\citep{masry2025chartgemma}, pretrained on chart corpora, reaches 27.5 (bar) and 22.9 (pie), on par with the off-the-shelf InternVL baseline (24.0, 26.0) and below skill-tuned InternVL (35.6, 40.6). Granite Vision 4.1-4b, fine-tuned on ChartNet for chart, table, and key-value extraction~\cite{kondic2026chartnet}, reaches only 17.5 and 18.7 on complex reasoning, below even the general-purpose baselines. Neither result closes the gap our skill supervision does: training on chart data does not supply the simple-skill competence that complex reasoning draws on. Both reference models are smaller than our models and trained under different regimes, so we read the comparison as situating rather than head-to-head.

\begin{table}[t]
\centering
\caption{Complex-reasoning accuracy (fuzzy \%) on the held-out CR-500 sets (bar / pie, 500 questions each). The blocks are \emph{not} directly comparable. Top block: our all-skill (\emph{P+G+SR}) ChartProbe fine-tune (mean over seeds 42/43/44); Bottom block: chart-oriented reference models evaluated zero-shot, shown for context, and both reference models are smaller than our models. Bold marks the best row per chart type.}
\label{tab:cr_ref}
\setlength{\tabcolsep}{6pt}\footnotesize
\begin{tabular}{@{}l cc@{}}
\toprule
& Bar & Pie \\
\midrule
\rowcolor{cprow}\multicolumn{3}{@{}l}{\textit{ChartProbe \emph{P+G+SR}}}\\
InternVL-3.5-8B    & \textbf{35.6} & \textbf{40.6} \\
Qwen3.5-9B      & 24.7 & 32.3 \\
LLaVA-1.6          & 26.2 & 24.2 \\
\midrule
\rowcolor{cprow}\multicolumn{3}{@{}l}{\textit{Chart-oriented reference models}}\\
ChartGemma (zero-shot)      & 27.5 & 22.9 \\
Granite Vision 4.1 (zero-shot)  & 17.5 & 18.7 \\
\bottomrule
\end{tabular}
\end{table}
\section{Discussion}
\label{sec:discussion}

Across the models and settings we study, much of what is scored as a complex-failures can be mitigated by up to 27.6\% gains by supervising the simple skills that reasoning draws on. Four observations support it. Supervising perception or grounding alone improves the simple-reasoning probe, and out of distribution it can improve it more than supervising simple reasoning directly. On held-out complex reasoning, every form of simple-skill supervision produces gains, with no complex reasoning questions seen at training time. We observe similar patterns in the out-of-distribution ChartQA dataset. A model can be brought to answer compositional questions better, in effect, by enhancing individual atomic skills.

\paragraph{Why grounding deserves its own skill.}
The common perception-versus-reasoning view in the literature attributes a binding error, where the model reads every value correctly but associates the named category with the wrong element, to reasoning. Our decomposition shows that this is a distinct and separately repairable error. Grounding supervision produces the largest gain in the grounding column (Table~\ref{tab:chartprobe_finetune}; $+19.5$ for InternVL on bar charts), a level that neither perception nor reasoning supervision alone reaches. This shows that grounding must be trained as such rather than acquired as a by-product of the other skills. We also observed that trade-offs exist, for example grounding-only supervision can affect other skills (for InternVL, $-17.9$ on simple reasoning). Hence, we have also provided scenarios of combined atomic skill supervision. 

\paragraph{Atomic P, G, SR skills transfer beyond charts.}
The clearest evidence comes from CLEVR, a non-chart domain absent from all fine-tuning. Supervising perception on bar charts raises Qwen's CLEVR
simple-reasoning probe from 21.6 to 79.5, above the 63.2 that direct
simple-reasoning supervision reaches on the same probe; for this model,
supervising perception alone helps reasoning more than supervising
reasoning does. Because the gain transfers to a domain with no bars, axes, or chart structure, it cannot be a chart-specific artifact.


\paragraph{Where the effect fails.}
The pattern is not uniform, and the exceptions are informative. In distribution, single-skill supervision can reduce another skill: for InternVL on bar, simple reasoning drops $-4.5$ and $-17.9$ after perception- and grounding-only fine-tuning. This is a sign that narrow supervision specializes a model at the expense of the capability we want to lift, and the combined configurations do not show it. More fundamentally, the effect presupposes a model whose earlier skills are trainable to a useful level. LLaVA's near-zero chart perception at baseline reflects a model that cannot deliver a usable read at all, and its complex-reasoning gains on charts are correspondingly the smallest we observe. Teaching perception only helps reasoning when there is something to teach: the model must be able to reach a usable level of perception in the first place.

\paragraph{Limitations.}
Our setting is deliberately controlled: single-series bar and pie charts and CLEVR scenes, where each skill has an exact ground-truth from the generating code. This is what makes the skills separable, and also what confines the claims to such domains. Extending the framework to natural images, where the skills blur and exact ground truth is unavailable, remains open. Our templates do not generalize to real images, and because the probe questions are template-generated, the trained skills correspond to those the templates define, though the complex-reasoning and ChartQA evaluations are authored independently. The combined configurations also use more training data than the single-skill ones. We do not compare against direct supervision on complex reasoning as a competing method, and we make no efficiency claim relative to it; our claim is diagnostic. Finally, our results are population-level: we identify which skill is limiting for a model and domain, not which skill produced a given wrong answer, and we do not explain at the representational level what keeps the earlier skills weak.

\section{Conclusion}
\label{sec:conclusion}

We introduced ChartProbe, a diagnostic that decomposes chart question answering into perception, grounding, and simple reasoning, each with an exact answer computed from the chart's generating code, and that targets each skill by intervention rather than by scoring. Under skill-targeted fine-tuning, supervising simple skills alone lifts held-out complex reasoning across three models and three domains, with no complex question ever in training. The benefit transfers to an unseen chart type, a non-chart domain, and an independently authored benchmark, while chart specialization, whether by pretraining or extraction-specific fine-tuning, does not reach it. The effect requires a model whose perception can already be trained to a useful level. Our diagnosis is also population-level: it identifies which skill is limiting for a model and domain, not which skill produced a given wrong answer. The practical takeaway is direct: before investing in
reasoning components, confirm that the model has perceived and bound the evidence it
must reason over. We release ChartProbe to support that check; a mechanistic account of why these skills remain weak is a direction for future work.
\bibliography{aaai2027}

@article{wei2022chain,
  title={Chain-of-thought prompting elicits reasoning in large language models},
  author={Wei, Jason and Wang, Xuezhi and Schuurmans, Dale and Bosma, Maarten and Xia, Fei and Chi, Ed and Le, Quoc V and Zhou, Denny and others},
  journal={Advances in neural information processing systems},
  volume={35},
  pages={24824--24837},
  year={2022}
}

@article{zhang2023multimodal,
  title={Multimodal chain-of-thought reasoning in language models},
  author={Zhang, Zhuosheng and Zhang, Aston and Li, Mu and Zhao, Hai and Karypis, George and Smola, Alex},
  journal={arXiv preprint arXiv:2302.00923},
  year={2023}
}

@article{chen2025g1,
  title={G1: Bootstrapping perception and reasoning abilities of vision-language model via reinforcement learning},
  author={Chen, Liang and Gao, Hongcheng and Liu, Tianyu and Huang, Zhiqi and Sung, Flood and Zhou, Xinyu and Wu, Yuxin and Chang, Baobao},
  journal={arXiv preprint arXiv:2505.13426},
  year={2025}
}

@inproceedings{yang2025r1,
  title={R1-onevision: Advancing generalized multimodal reasoning through cross-modal formalization},
  author={Yang, Yi and He, Xiaoxuan and Pan, Hongkun and Jiang, Xiyan and Deng, Yan and Yang, Xingtao and Lu, Haoyu and Yin, Dacheng and Rao, Fengyun and Zhu, Minfeng and others},
  booktitle={Proceedings of the IEEE/CVF International Conference on Computer Vision},
  pages={2376--2385},
  year={2025}
}

@article{huang2025vision,
  title={Vision-r1: Incentivizing reasoning capability in multimodal large language models},
  author={Huang, Wenxuan and Jia, Bohan and Zhai, Zijie and Cao, Shaosheng and Ye, Zheyu and Zhao, Fei and Xu, Zhe and Tang, Xu and Hu, Yao and Lin, Shaohui},
  journal={arXiv preprint arXiv:2503.06749},
  year={2025}
}

@article{xiao2025perception,
  title={Perception-R1: Advancing Multimodal Reasoning Capabilities of MLLMs via Visual Perception Reward},
  author={Xiao, Tong and Xu, Xin and Huang, Zhenya and Gao, Hongyu and Liu, Quan and Liu, Qi and Chen, Enhong},
  journal={arXiv preprint arXiv:2506.07218},
  year={2025}
}

@article{yu2026perception,
  title={Perception-r1: Pioneering perception policy with reinforcement learning},
  author={Yu, En and Lin, Kangheng and Zhao, Liang and Wei, Yana and Peng, Yuang and Wei, Haoran and Sun, Jianjian and Han, Chunrui and Ge, Zheng and Zhang, Xiangyu and others},
  journal={Advances in Neural Information Processing Systems},
  volume={38},
  pages={94827--94853},
  year={2026}
}

@article{wu2025vtool,
  title={Vtool-r1: Vlms learn to think with images via reinforcement learning on multimodal tool use},
  author={Wu, Mingyuan and Yang, Jingcheng and Jiang, Jize and Li, Meitang and Yan, Kaizhuo and Yu, Hanchao and Zhang, Minjia and Zhai, Chengxiang and Nahrstedt, Klara},
  journal={arXiv preprint arXiv:2505.19255},
  year={2025}
}

@article{alain2016understanding,
  title={Understanding intermediate layers using linear classifier probes},
  author={Alain, Guillaume and Bengio, Yoshua},
  journal={arXiv preprint arXiv:1610.01644},
  year={2016}
}

@inproceedings{tenney2019bert,
  title={BERT rediscovers the classical NLP pipeline},
  author={Tenney, Ian and Das, Dipanjan and Pavlick, Ellie},
  booktitle={Proceedings of the 57th annual meeting of the association for computational linguistics},
  pages={4593--4601},
  year={2019}
}

@inproceedings{selvaraju2020squinting,
  title={Squinting at vqa models: Introspecting vqa models with sub-questions},
  author={Selvaraju, Ramprasaath R and Tendulkar, Purva and Parikh, Devi and Horvitz, Eric and Ribeiro, Marco Tulio and Nushi, Besmira and Kamar, Ece},
  booktitle={Proceedings of the IEEE/CVF Conference on Computer Vision and Pattern Recognition},
  pages={10003--10011},
  year={2020}
}

@article{yi2018neural,
  title={Neural-symbolic vqa: Disentangling reasoning from vision and language understanding},
  author={Yi, Kexin and Wu, Jiajun and Gan, Chuang and Torralba, Antonio and Kohli, Pushmeet and Tenenbaum, Josh},
  journal={Advances in neural information processing systems},
  volume={31},
  year={2018}
}

@inproceedings{jin2026seeing,
  title={Seeing Without Understanding: Disentangling Perception, Reasoning, and Simulation in VLM Gameplay},
  author={Jin, Dingyang and He, Jiawei and Lo, Calvin and Hu, Steven and Rad, Ryan},
  booktitle={Proceedings of the 43rd International Conference on Machine Learning (ICML)},
  year={2026}
}

@article{li2025self,
  title={Self-rewarding vision-language model via reasoning decomposition},
  author={Li, Zongxia and Yu, Wenhao and Huang, Chengsong and Liang, Zhenwen and Liu, Rui and Liu, Fuxiao and Che, Jingxi and Yu, Dian and Boyd-Graber, Jordan and Mi, Haitao and others},
  journal={arXiv preprint arXiv:2508.19652},
  year={2025}
}

@inproceedings{lee2026visdot,
  title={VisDoT: Enhancing Visual Reasoning through Human-Like Interpretation Grounding and Decomposition of Thought},
  author={Lee, Eunsoo and Lee, Jeongwoo and Hong, Minki and Choi, Jangho and Kim, Jihie},
  booktitle={Findings of the Association for Computational Linguistics: EACL 2026},
  pages={610--640},
  year={2026}
}

@inproceedings{masry2022chartqa,
  title={Chartqa: A benchmark for question answering about charts with visual and logical reasoning},
  author={Masry, Ahmed and Tan, Jia Qing and Joty, Shafiq and Hoque, Enamul and others},
  booktitle={Findings of the association for computational linguistics: ACL 2022},
  pages={2263--2279},
  year={2022}
}

@inproceedings{methani2020plotqa,
  title={Plotqa: Reasoning over scientific plots},
  author={Methani, Nitesh and Ganguly, Pritha and Khapra, Mitesh M and Kumar, Pratyush},
  booktitle={Proceedings of the ieee/cvf winter conference on applications of computer vision},
  pages={1527--1536},
  year={2020}
}

@inproceedings{johnson2017clevr,
  title={Clevr: A diagnostic dataset for compositional language and elementary visual reasoning},
  author={Johnson, Justin and Hariharan, Bharath and Van Der Maaten, Laurens and Fei-Fei, Li and Lawrence Zitnick, C and Girshick, Ross},
  booktitle={Proceedings of the IEEE conference on computer vision and pattern recognition},
  pages={2901--2910},
  year={2017}
}

@inproceedings{masry2025chartgemma,
  title={Chartgemma: Visual instruction-tuning for chart reasoning in the wild},
  author={Masry, Ahmed and Thakkar, Megh and Bajaj, Aayush and Kartha, Aaryaman and Hoque, Enamul and Joty, Shafiq},
  booktitle={Proceedings of the 31st International Conference on Computational Linguistics: Industry Track},
  pages={625--643},
  year={2025}
}

@article{han2023chartllama,
  title={Chartllama: A multimodal llm for chart understanding and generation},
  author={Han, Yucheng and Zhang, Chi and Chen, Xin and Yang, Xu and Wang, Zhibin and Yu, Gang and Fu, Bin and Zhang, Hanwang},
  journal={arXiv preprint arXiv:2311.16483},
  year={2023}
}

@inproceedings{liu2023matcha,
  title={Matcha: Enhancing visual language pretraining with math reasoning and chart derendering},
  author={Liu, Fangyu and Piccinno, Francesco and Krichene, Syrine and Pang, Chenxi and Lee, Kenton and Joshi, Mandar and Altun, Yasemin and Collier, Nigel and Eisenschlos, Julian},
  booktitle={Proceedings of the 61st Annual Meeting of the Association for Computational Linguistics (Volume 1: Long Papers)},
  pages={12756--12770},
  year={2023}
}

@inproceedings{kondic2026chartnet,
  title={Chartnet: A million-scale, high-quality multimodal dataset for robust chart understanding},
  author={Kondic, Jovana and Li, Pengyuan and Joshi, Dhiraj and Sanchez, Isaac and Wiesel, Ben and Abedin, Shafiq and Alfassy, Amit and Schwartz, Eli and Caraballo, Daniel and Cinar, Yagmur Gizem and others},
  booktitle={Proceedings of the IEEE/CVF Conference on Computer Vision and Pattern Recognition},
  pages={15922--15932},
  year={2026}
}

@inproceedings{liu2023deplot,
  title={DePlot: One-shot visual language reasoning by plot-to-table translation},
  author={Liu, Fangyu and Eisenschlos, Julian and Piccinno, Francesco and Krichene, Syrine and Pang, Chenxi and Lee, Kenton and Joshi, Mandar and Chen, Wenhu and Collier, Nigel and Altun, Yasemin},
  booktitle={Findings of the Association for Computational Linguistics: ACL 2023},
  pages={10381--10399},
  year={2023}
}

@inproceedings{tong2024eyes,
  title={Eyes wide shut? exploring the visual shortcomings of multimodal llms},
  author={Tong, Shengbang and Liu, Zhuang and Zhai, Yuexiang and Ma, Yi and LeCun, Yann and Xie, Saining},
  booktitle={Proceedings of the IEEE/CVF conference on computer vision and pattern recognition},
  pages={9568--9578},
  year={2024}
}

@article{wang2024charxiv,
  title={Charxiv: Charting gaps in realistic chart understanding in multimodal llms},
  author={Wang, Zirui and Xia, Mengzhou and He, Luxi and Chen, Howard and Liu, Yitao and Zhu, Richard and Liang, Kaiqu and Wu, Xindi and Liu, Haotian and Malladi, Sadhika and others},
  journal={Advances in Neural Information Processing Systems},
  volume={37},
  pages={113569--113697},
  year={2024}
}

@article{rahmanzadehgervi2024vision,
  title={Vision language models are blind: Failing to translate detailed visual features into words},
  author={Rahmanzadehgervi, Pooyan and Bolton, Logan and Taesiri, Mohammad Reza and Nguyen, Anh Totti},
  journal={arXiv preprint arXiv:2407.06581},
  year={2024}
}

@article{gu2025composition,
  title={Composition-Grounded Instruction Synthesis for Visual Reasoning},
  author={Gu, Xinyi and Mao, Jiayuan and Hong, Zhang-Wei and Yu, Zhuoran and Li, Pengyuan and Joshi, Dhiraj and Feris, Rogerio and He, Zexue},
  journal={arXiv preprint arXiv:2510.15040},
  year={2025}
}

@inproceedings{masry2025chartqapro,
  title={Chartqapro: A more diverse and challenging benchmark for chart question answering},
  author={Masry, Ahmed and Islam, Mohammed Saidul and Ahmed, Mahir and Bajaj, Aayush and Kabir, Firoz and Kartha, Aaryaman and Laskar, Md Tahmid Rahman and Rahman, Mizanur and Rahman, Shadikur and Shahmohammadi, Mehrad and others},
  booktitle={Findings of the Association for Computational Linguistics: ACL 2025},
  pages={19123--19151},
  year={2025}
}

@article{kahou2017figureqa,
  title={Figureqa: An annotated figure dataset for visual reasoning},
  author={Kahou, Samira Ebrahimi and Michalski, Vincent and Atkinson, Adam and K{\'a}d{\'a}r, {\'A}kos and Trischler, Adam and Bengio, Yoshua},
  journal={arXiv preprint arXiv:1710.07300},
  year={2017}
}

@inproceedings{kafle2018dvqa,
  title={Dvqa: Understanding data visualizations via question answering},
  author={Kafle, Kushal and Price, Brian and Cohen, Scott and Kanan, Christopher},
  booktitle={Proceedings of the IEEE conference on computer vision and pattern recognition},
  pages={5648--5656},
  year={2018}
}

@article{wang2025internvl3,
  title={Internvl3. 5: Advancing open-source multimodal models in versatility, reasoning, and efficiency},
  author={Wang, Weiyun and Gao, Zhangwei and Gu, Lixin and Pu, Hengjun and Cui, Long and Wei, Xingguang and Liu, Zhaoyang and Jing, Linglin and Ye, Shenglong and Shao, Jie and others},
  journal={arXiv preprint arXiv:2508.18265},
  year={2025}
}

@article{team2026qwen3,
  title={Qwen3. 5-omni technical report},
  author={Team, Qwen},
  journal={arXiv preprint arXiv:2604.15804},
  year={2026}
}

@misc{liu2023improvedllava,
          author={Liu, Haotian and Li, Chunyuan and Li, Yuheng and Lee, Yong Jae},
          title={Improved Baselines with Visual Instruction Tuning}, 
          publisher={arXiv:2310.03744},
          year={2023},
  }

@misc{rapidfuzz,
  author       = {Max Bachmann},
  title        = {{RapidFuzz}: Rapid Fuzzy String Matching in {Python}},
  year         = {2024},
  howpublished = {\url{https://rapidfuzz.github.io/RapidFuzz/}},
  note         = {Version X.Y.Z}
}
\clearpage 
\definecolor{cpink}{HTML}{262B33}   
\definecolor{cpmuted}{HTML}{6B7280} 
\definecolor{cprule}{HTML}{E1E7EE}  
\definecolor{cpP}{HTML}{2E8B7F}     
\definecolor{cpG}{HTML}{C6803B}     
\definecolor{cpR}{HTML}{3E6DA6}     
\definecolor{cpsoftP}{HTML}{F4FBF9}  
\definecolor{cpsoftG}{HTML}{FDF4EA}  
\definecolor{cpsoftR}{HTML}{F3F8FD}  
\definecolor{cpcodebg}{HTML}{F6F8FA} 
\definecolor{cpslotc}{HTML}{D2691E}  
\definecolor{cpok}{HTML}{1F8A70}     
\definecolor{cpfail}{HTML}{C0392B}   

\captionsetup{font=small,labelfont=bf,skip=6pt}

\lstdefinestyle{cppystyle}{
  language=Python,
  basicstyle=\ttfamily\footnotesize\color{cpink},
  keywordstyle=\bfseries\color{cpR},
  commentstyle=\itshape\color{cpmuted},
  stringstyle=\color{cpP},
  backgroundcolor=\color{cpcodebg},
  frame=single, rulecolor=\color{cprule},
  framesep=4pt, xleftmargin=4pt, xrightmargin=4pt,
  breaklines=true, showstringspaces=false, tabsize=4,
  morekeywords={sns,plt,figsize,palette},
}
\lstdefinestyle{cpcsvstyle}{
  basicstyle=\ttfamily\footnotesize\color{cpink},
  backgroundcolor=\color{cpcodebg},
  frame=single, rulecolor=\color{cprule},
  framesep=4pt, xleftmargin=4pt, xrightmargin=4pt,
  breaklines=true,
}

\providecommand{\cptid}{}\renewcommand{\cptid}[1]{{\footnotesize\texttt{#1}}}
\providecommand{\cptmpl}{}\renewcommand{\cptmpl}[1]{{\footnotesize\texttt{[#1]}}}
\providecommand{\cpslot}{}\renewcommand{\cpslot}[1]{\textcolor{cpslotc}{\texttt{\{#1\}}}}
\providecommand{\cpfilled}{}\renewcommand{\cpfilled}[1]{\textcolor{cpok}{\textbf{#1}}}
\providecommand{\cpbad}{}\renewcommand{\cpbad}[1]{\textcolor{cpfail}{#1}}
\providecommand{\cpgood}{}\renewcommand{\cpgood}[1]{\textcolor{cpok}{#1}}
\providecommand{\cpfmt}{}\renewcommand{\cpfmt}[1]{\textcolor{cpmuted}{#1}}

\providecommand{\cppok}{}\renewcommand{\cppok}{\textcolor{cpok}{\textbf{ok}}}
\providecommand{\cppflip}{}\renewcommand{\cppflip}{\textcolor{cpok}{\textbf{fixed}}}
\providecommand{\cppmiss}{}\renewcommand{\cppmiss}{\textcolor{cpfail}{\textbf{miss}}}
\providecommand{\cppregr}{}\renewcommand{\cppregr}{\textcolor{cpfail}{\textbf{regress}}}

\providecommand{\cpstagetag}{}\renewcommand{\cpstagetag}[2]{%
  \tikz[baseline=-0.6ex]{%
    \node[fill=#1, rounded corners=2.5pt, inner xsep=5pt, inner ysep=2pt]
      {\textcolor{white}{\scriptsize\bfseries #2}};}%
}

\providecommand{\cpstagehead}{}\renewcommand{\cpstagehead}[2]{%
  \par\smallskip\noindent
  \tikz{\node[fill=#1, rounded corners=3pt, inner xsep=9pt, inner ysep=3.5pt]
    {\textcolor{white}{\small\bfseries #2}};}%
  \par\nopagebreak\smallskip
}

\providecommand{\cpsafeimg}{}\renewcommand{\cpsafeimg}[2]{%
  \IfFileExists{#1}{\includegraphics[width=#2]{#1}}%
  {\fbox{\parbox[c][2.2cm][c]{#2}{\centering\scriptsize\ttfamily missing:\\ #1}}}%
}

\providecommand{\cpbox}{}\renewcommand{\cpbox}[2]{%
  \par\smallskip\noindent
  \begin{tikzpicture}
    \node[draw=cprule, line width=0.5pt, rounded corners=3pt, fill=#1,
          inner xsep=12pt, inner ysep=8pt,
          text width=\dimexpr\linewidth-24pt\relax, align=left] (cpb) {\small #2};
  \end{tikzpicture}%
  \par\smallskip
}
\providecommand{\cpcallout}{}\renewcommand{\cpcallout}[1]{\cpbox{cpcodebg}{#1}}

\section{ \LARGE Supplementary Materials}
\vspace{1em}

This supplementary material provides additional analyses, walkthroughs, and results that complement the main paper. We describe how skill-targeted probes are generated, walk through an illustrative example of ChartProbe end to end, present qualitative traces of model behavior, and report a volume-matched ablation.

\section{Generating skill-targeted Probes}

\label{sec:supp-extraction}

Given the underlying table (derived from the generation code or CSV), ChartProbe instantiates a fixed template bank and deterministically computes the corresponding ground-truth answers from the table, with element bounding boxes additionally used for grounding tasks. 

Figure~\ref{fig:data_generation} shows the full pipeline: each chart originates from executable code that emits a table, which is both rendered into the image and used to bind template slots, so every probe question carries programmatic ground truth by construction. The rendered chart and the instantiated question are then paired as input to the VLM under evaluation.


\begin{center}
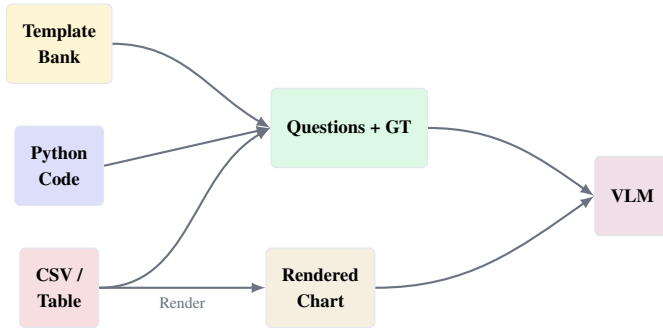

\begin{tikzpicture}[
  node distance=0.55cm and 0.55cm,
  box/.style={draw=cprule, rounded corners=2pt, align=center,
    font=\scriptsize\bfseries, inner xsep=6pt, inner ysep=5pt,
    minimum height=1.05cm},
  arr/.style={-latex, thick, color=cpmuted},
  lbl/.style={font=\tiny, text=cpmuted}
]
\node[box, fill=cpcodebg!90!blue] (code) {Python\\Code};
\node[box, fill=cpcodebg!90!red, below=of code] (csv) {CSV /\\Table};
\node[box, fill=cpsoftG!90!yellow, above=of code] (tpl) {Template\\Bank};

\node[box, fill=cpsoftR!90!green, right=2.2cm of code, yshift=0.5cm] (qa) {Questions + GT};

\node[box, fill=cpsoftP!90!orange, right=2.2cm of csv] (chart) {Rendered\\Chart};

\node[box, fill=cpcodebg!90!purple, right=2.2cm of qa, yshift=-0.9cm] (vlm) {VLM};

\draw[arr] (code.east) -- (qa.west);
\draw[arr] (tpl.east) to[out=0, in=150] (qa.west);
\draw[arr] (csv.east) to[out=0, in=210] (qa.west);

\draw[arr] (csv.east) -- node[lbl, below]{Render} (chart.west);

\draw[arr] (qa.east) to[out=0, in=145] (vlm.west);
\draw[arr] (chart.east) to[out=0, in=215] (vlm.west);
\end{tikzpicture}
\captionof{figure}{Python code, the CSV/table, and the
  template bank together generate each probe question with its programmatic ground
  truth; the same table is rendered into a chart. The rendered chart and the
  instantiated question are the two inputs fed to the VLM.}
\label{fig:data_generation}
\end{center}

\paragraph{Answer matching.} All questions use the evaluation protocol of the main paper: binary, count, and label answers by exact match after lowercasing and whitespace stripping, numeric answers within $\pm5\%$ relative tolerance, and list answers by recall against the ground truth. Answers are unit-insensitive, so \texttt{3.1} and \texttt{3.1\%} both score correct, and label answers are matched under a fixed alias table, so \texttt{UK} and \texttt{United Kingdom} are equivalent. Bounding boxes are scored by IoU at a $0.5$ threshold, with the achieved IoU shown in parentheses.

\paragraph{Template identifiers.} Template IDs name the canonical form and are orientation-adapted at instantiation: \cptid{G-VT-LEFTMOST} resolves to the topmost bar on a horizontal chart and to the largest segment on a pie, so the surface question may not repeat the word in the identifier.

\subsection{Template Bank}
Table~\ref{tab:template_bank} lists the complete ChartProbe template bank in
its bar-chart form: 26 schemas across the three atomic skills (8 perception, 9
grounding, 9 simple reasoning). Each row pairs a question schema with an answer
function over the table. Instantiating a schema binds its braced slots (a label
column, a value column, two named elements, or the chart's extrema) to that
chart's actual headers and values, then evaluates the answer function on the
underlying data; every GT answer is thus computed from the chart's generating
code rather than read from the image, making it exact and annotation-free. The
\emph{Type} column gives the scoring rule for each answer. Crucially, the three
skills differ in what a schema may depend on, not merely in difficulty.
Perception schemas resolve from bar geometry alone (counts, relative heights,
extremal position, left-to-right trend) and never require reading an axis value
or naming a category, which is why they carry an instruction to ignore axis
text. Grounding schemas bind a language label to a visual element in either
direction, including reading a named bar's value and predicting its bounding
box. Simple-reasoning schemas apply a single operation (rank, sum, mean,
difference, or percentage) over the values those elements carry. Where two
skills touch the same quantity they still differ in demand: a perception schema
asks how many bars sit above the average height as a holistic visual estimate
with no value computed (\texttt{P-COUNT-ABOVE-AVG}), whereas the corresponding
simple-reasoning schema asks which named categories exceed the computed average
(\texttt{R-RANK-ABOVE-AVG}), requiring the actual values to be recovered and
compared.

The pie bank reuses these 26 schemas unchanged, since a pie chart carries the
same label-to-value mapping as a bar chart and differs only in geometry.
Instantiation re-resolves the geometric references: positional slots
(\emph{leftmost}, \emph{rightmost}) bind to segment order read clockwise from
twelve o'clock, \emph{tallest}/\emph{shortest} bind to the largest and smallest
segment by arc, above-average comparisons bind to share of the whole, and the
bounding-box target binds to a segment's wedge region. The answer functions are
untouched, since they operate on the same underlying label-value pairs, so GT is
still computed from the generating code.

CLEVR follows the same diagnostic design principle as ChartProbe, while adapting the schema definitions to its scene-graph representation. CLEVR does not reuse the chart schemas because its scenes contain no axes, labels, or tabular values. Instead, each skill is defined over scene-graph primitives (object color, shape, material, size, and spatial relations), with ground truth computed directly from the underlying scene graph. Perception schemas count objects or attributes, grounding schemas localize or identify objects and their attributes or relations, and simple-reasoning schemas perform a single operation such as comparing counts, determining the majority material, or testing whether two objects share a material. As with charts, every answer is derived automatically from the generating representation, requiring no manual annotation.

\begin{table*}[t]
\centering
\caption{\textbf{The complete ChartProbe template bank} (bar-chart form):
26 templates spanning 8 perception, 9 grounding, and 9 simple-reasoning
schemas. Each template pairs a natural-language question schema with an
answer function; instantiating it on a chart binds the braced slots to that
chart's column headers and extrema and evaluates the answer from the
generating code, so every GT answer is exact and requires no annotation.
Perception schemas are answerable from bar geometry alone, with no value read
from the axis; grounding schemas bind a label to an element (or read a named
bar's value); simple-reasoning schemas apply a single operation over the
bound values. The pie bank reuses these 26 schemas under re-resolved geometry
(e.g.\ \emph{leftmost}/\emph{tallest} resolve to the first/largest segment); CLEVR uses a separate scene-graph schema set, released in code.
The \emph{Type} column names the scoring rule (Section~\ref{subsec:eval}).}
\label{tab:template_bank}
\setlength{\tabcolsep}{6pt}
\renewcommand{\arraystretch}{1.15}
\small
\begin{tabularx}{\textwidth}{@{}l >{\raggedright\arraybackslash}X l@{}}
\toprule
\textbf{ID} & \textbf{Question schema} & \textbf{Type} \\
\midrule
\rowcolor{cprow}\multicolumn{3}{@{}l}{\textbf{Perception}\quad\textit{(8; geometry only, no axis read)}} \\
\texttt{P-COUNT-TOTAL}     & How many bars are in the chart? & count \\
\texttt{P-COUNT-ABOVE-AVG} & How many bars appear to be above the average height? & count \\
\texttt{P-HEIGHT-2X}       & Is the tallest bar more than twice the height of the shortest bar? & binary \\
\texttt{P-HEIGHT-LR}       & Is the leftmost bar taller than the rightmost bar? & binary \\
\texttt{P-HEIGHT-MIDDLE}   & How many bars are taller than the middle bar? & count \\
\texttt{P-POS-TALLEST}     & Is the tallest bar on the left side or the right side of the chart? & label \\
\texttt{P-TREND-LR}        & Do the bars increase in height from left to right? & binary \\
\texttt{P-TREND-PATTERN}   & What is the overall trend of the bars from left to right? & label \\
\addlinespace[2pt]
\rowcolor{cprow}\multicolumn{3}{@{}l}{\textbf{Grounding}\quad\textit{(9; label--element binding, value, and box)}} \\
\texttt{G-VT-TALLEST}   & What is the name of the tallest bar? & label \\
\texttt{G-VT-SHORTEST}  & Which \cpslot{label\_col} does the shortest bar represent? & label \\
\texttt{G-VT-LEFTMOST}  & What is the \cpslot{label\_col} of the leftmost bar? & label \\
\texttt{G-VT-RIGHTMOST} & What is the \cpslot{label\_col} of the rightmost bar? & label \\
\texttt{G-TV-VALUE-A}   & What is the \cpslot{value\_col} for \cpslot{label\_a}? & value \\
\texttt{G-TV-VALUE-B}   & What is the \cpslot{value\_col} for \cpslot{label\_b}? & value \\
\texttt{G-TV-COMPARE}   & Is the bar for \cpslot{label\_a} taller than the bar for \cpslot{label\_b}? & binary \\
\texttt{G-TV-ABOVE-N}   & Is the bar for \cpslot{label} above the value of \cpslot{n} on the y-axis? & binary \\
\texttt{G-BBOX-ELEMENT} & What is the bounding box of the bar for \cpslot{label}? & bbox \\
\addlinespace[2pt]
\rowcolor{cprow}\multicolumn{3}{@{}l}{\textbf{Simple reasoning}\quad\textit{(9; one operation over bound values)}} \\
\texttt{R-RANK-TOP3}       & What are the top 3 \cpslot{label\_col}s by \cpslot{value\_col}? & list \\
\texttt{R-RANK-HIGHER}     & Which has a higher \cpslot{value\_col}: \cpslot{label\_a} or \cpslot{label\_b}? & label \\
\texttt{R-RANK-ABOVE-AVG}  & Which \cpslot{label\_col}s have a \cpslot{value\_col} above the average? & list \\
\texttt{R-ARITH-TOTAL}     & What is the total \cpslot{value\_col} across all bars combined? & value \\
\texttt{R-ARITH-MEAN}      & What is the average \cpslot{value\_col} across all \cpslot{label\_col}s? & value \\
\texttt{R-ARITH-DIFF}      & What is the difference in \cpslot{value\_col} between \cpslot{label\_a} and \cpslot{label\_b}? & value \\
\texttt{R-ARITH-PCT}       & What percentage of the total \cpslot{value\_col} does \cpslot{label} represent? & value \\
\texttt{R-MULTI-DIFF-MEAN} & How much higher is \cpslot{top\_label} compared to the average \cpslot{value\_col}? & value \\
\texttt{R-MULTI-CLOSEST}   & Which two \cpslot{label\_col}s are closest in \cpslot{value\_col}? & list \\
\bottomrule
\end{tabularx}
\end{table*}

\section{Illustrative Walkthrough of ChartProbe}
\label{subsec:bar-chart-example}

Starting from the source code and its underlying
values, we draw templates from the per-domain pool to produce nine probes,
three targeting each skill: perception probes ask about raw visual
attributes (bar heights, colors); grounding probes tie a queried element to its position or extent; and
simple-reasoning probes pose a single-step readout (such as a count or
comparison) over the chart. 

\begin{figure}[ht]
\centering
\cpsafeimg{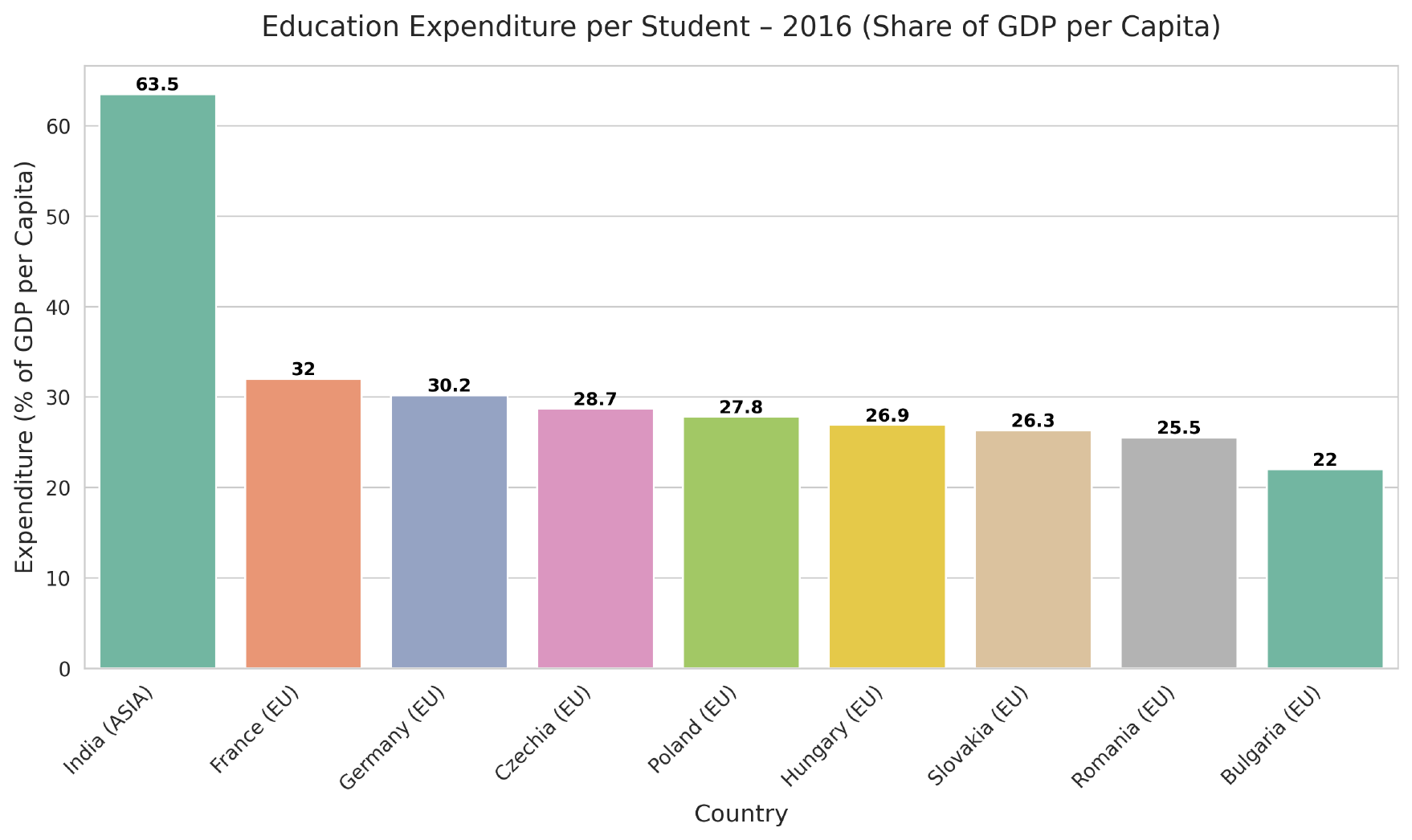}{0.34\textwidth}
\caption{Rendered bar chart for the shown example, the image input passed to the VLM alongside the instantiated probe questions using ChartProbe}
\label{fig:rendered-chart}
\end{figure}

\subsection{1- Source Code}

\begin{lstlisting}[style=cppystyle,
    basicstyle=\footnotesize\ttfamily,
    breaklines=true,
    breakatwhitespace=true,
    columns=fullflexible,
    keepspaces=true]
countries = [
    "Bulgaria (EU)", "India (ASIA)",
    "Slovakia (EU)", "Poland (EU)",
    "Czechia (EU)", "Romania (EU)",
    "Hungary (EU)", "Germany (EU)",
    "France (EU)",
]
expenditure_2016 = [
    22.0, 63.5, 26.3, 27.8, 28.7,
    25.5, 26.9, 30.2, 32.0,
]
order = np.argsort(expenditure_2016)[::-1]
sns.barplot(
    x=[countries[i] for i in order],
    y=[expenditure_2016[i] for i in order],
    palette="Set2",
)
\end{lstlisting}

\subsection{2- Table (CSV)}
\begin{lstlisting}[style=cpcsvstyle]
Country,Expenditure (% of GDP per Capita)
India (ASIA),63.5
France (EU),32
Germany (EU),30.2
Czechia (EU),28.7
Poland (EU),27.8
Hungary (EU),26.9
Slovakia (EU),26.3
Romania (EU),25.5
Bulgaria (EU),22
\end{lstlisting}

\subsection{3- Template}
Here we discuss how the template can be used to automatically generate questions and have them linked to their corresponding ground-truth. Slots in \textcolor{cpslotc}{orange} are bound from the table; the filled
surface form is what the model sees. Because every slot is bound directly from the table values, each generated question carries an exact, automatically derived answer, with no manual annotation required.

\cpstagehead{cpP}{Perception}
\begin{enumerate}
\item \cptid{P-COUNT-ABOVE-AVG}%
  \cpbox{cpsoftP}{%
    Rule: count $\lvert\{v : v > \bar{v}\}\rvert$\quad\\
    \textbf{GT:} \cpfilled{2}
    {\footnotesize\color{cpmuted}($63.5,\ 32 > 31.43$)}}
\item \cptid{P-HEIGHT-2X}%
  \cpbox{cpsoftP}{%
    Rule: Yes iff $\max > 2\min$\quad\\
    \textbf{GT:} \cpfilled{Yes}
    {\footnotesize\color{cpmuted}($63.5 > 2\times22$)}}
\item \cptid{P-TREND-PATTERN}%
  \cpbox{cpsoftP}{%
    Rule: monotonicity of bars left$\rightarrow$right\quad\\
    \textbf{GT:} \cpfilled{Consistently decreasing}}
\end{enumerate}

\cpstagehead{cpG}{Grounding}
\begin{enumerate}
\item \cptid{G-BBOX-ELEMENT}%
  \cpbox{cpsoftG}{%
    Template: What is the bounding box of the bar for \cpslot{tallest\_label}?\\
    Filled: \ldots\ for \cpfilled{India (ASIA)}?\quad\\
    \textbf{GT:} \cpfilled{[76, 118, 158, 788]}}
\item \cptid{G-VT-LEFTMOST}%
  \cpbox{cpsoftG}{%
    Template: What is the \cpslot{label\_col} of the leftmost bar?\\
    Filled: What is the \cpfilled{Country} of the leftmost bar?\quad\\
    \textbf{GT:} \cpfilled{India (ASIA)}}
\item \cptid{G-TV-VALUE-B}%
  \cpbox{cpsoftG}{%
    Template: What is the \cpslot{value\_col} for \cpslot{second\_label}?\\
    Filled: \ldots\ \cpfilled{Expenditure (\% of GDP per Capita)} for \cpfilled{France (EU)}?\quad\\
    \textbf{GT:} \cpfilled{32}}
\end{enumerate}

\cpstagehead{cpR}{Simple Reasoning}
\begin{enumerate}
\item \cptid{R-RANK-TOP3}%
  \cpbox{cpsoftR}{%
    Template: What are the top 3 \cpslot{label\_col}s by \cpslot{value\_col}?\\
    Filled: \ldots\ top 3 \cpfilled{Country}s by \cpfilled{Expenditure\ldots}?\quad\\
    \textbf{GT:} \cpfilled{India (ASIA), France (EU), Germany (EU)}}
\item \cptid{R-ARITH-TOTAL}%
  \cpbox{cpsoftR}{%
    Template: What is the total \cpslot{value\_col} across all bars combined?\\
    Filled: \ldots\ total \cpfilled{Expenditure\ldots}?\quad\\
    \textbf{GT:} \cpfilled{282.9}}
\item \cptid{R-MULTI-DIFF-MEAN}%
  \cpbox{cpsoftR}{%
    Template: How much higher is \cpslot{top\_label} compared to the average \cpslot{value\_col}?\\
    Filled: \ldots\ \cpfilled{India (ASIA)} \ldots\ average \cpfilled{Expenditure\ldots}?\quad\\
    \textbf{GT:} \cpfilled{32.07}
    {\footnotesize\color{cpmuted}($63.5 - 31.43$)}}
\end{enumerate}

\begin{table}[t]
\centering
\caption{All nine instantiated ChartProbe questions.}
\label{tab:suppl-9q}
\scriptsize\setlength{\tabcolsep}{2.5pt}
\begin{tabularx}{\columnwidth}{@{}c >{\raggedright\arraybackslash}X >{\raggedright\arraybackslash}p{1.9cm}@{}}
\toprule
& Question & Answer \\
\midrule
\cpstagetag{cpP}{P}
  & How many bars appear to be above the average height?\newline\cptid{P-COUNT-ABOVE-AVG}
  & \textbf{2} \\
\cpstagetag{cpP}{P}
  & Is the tallest bar more than twice the height of the shortest bar?\newline\cptid{P-HEIGHT-2X}
  & \textbf{Yes} \\
\cpstagetag{cpP}{P}
  & What is the overall trend of the bars from left to right?\newline\cptid{P-TREND-PATTERN}
  & \textbf{Consistently decreasing} \\
\midrule
\cpstagetag{cpG}{G}
  & What is the bounding box of the bar for India (ASIA)?\newline\cptid{G-BBOX-ELEMENT}
  & \textbf{[76, 118, 158, 788]} \\
\cpstagetag{cpG}{G}
  & What is the Country of the leftmost bar?\newline\cptid{G-VT-LEFTMOST}
  & \textbf{India (ASIA)} \\
\cpstagetag{cpG}{G}
  & What is the Expenditure (\% of GDP per Capita) for France (EU)?\newline\cptid{G-TV-VALUE-B}
  & \textbf{32} \\
\midrule
\cpstagetag{cpR}{R}
  & What are the top 3 Countrys by Expenditure (\% of GDP per Capita)?\newline\cptid{R-RANK-TOP3}
  & \textbf{India (ASIA), France (EU), Germany (EU)} \\
\cpstagetag{cpR}{R}
  & What is the total Expenditure (\% of GDP per Capita) across all bars combined?\newline\cptid{R-ARITH-TOTAL}
  & \textbf{282.9} \\
\cpstagetag{cpR}{R}
  & How much higher is India (ASIA) compared to the average Expenditure (\% of GDP per Capita)?\newline\cptid{R-MULTI-DIFF-MEAN}
  & \textbf{32.07} \\
\bottomrule
\end{tabularx}
\end{table}

  The pipeline moves from the underlying code/CSV, which fixes the
  schema and values, through templates, which fix linguistic form and
  answer rules, to instantiation, which yields a reproducible $(Q, A)$
  pair for each of the nine probes. The nine instantiated questions for
  this sample are listed in Table~\ref{tab:suppl-9q}, and the chart
  itself is rendered in Figure~\ref{fig:rendered-chart}. These two, the
  rendered image and its instantiated questions, are what we feed to the
  VLM. \emph{complex-reasoning} questions are separate.

\section{Qualitative Traces}
\label{sec:supp-heroes}

We trace three samples question by question, comparing the off-the-shelf baseline against a perception-only fine-tune of the same model. These traces are illustrative, not evidentiary: their purpose is to show \emph{what a per-cell change looks like}, which aggregate accuracy hides, and to make plain why we do not rest any claim on a single trace. The rows move in every direction. Some are genuine repairs, such as the trend probe in the bar-chart trace below, where the baseline reports a steady decrease on a series that fluctuates and the fine-tune corrects it. Some are format repairs: the baseline has the right content but buries it in reasoning that the matcher scores wrong, and the fine-tune surfaces a scoreable answer. We show the mixed and negative cells rather than curate a clean story.

\subsection{Bar chart sample}
\label{sec:hero-bar}

The baseline handles value lookup and ranking on this chart but misreads the overall shape of the series, reporting a consistent decrease where the values fluctuate ($4.5 \rightarrow 2.7 \rightarrow 3.1 \rightarrow 2.4$). After Perception-only fine-tuning the trend probe flips to correct. The complex-reasoning question also flips. Table~\ref{tab:trace-bar-hero} traces every probe on this chart.

\subsection{Pie chart sample}
\label{sec:hero-pie}

This chart isolates the leaked chain-of-thought failure mode: the baseline emits step-by-step reasoning on several questions and never commits to a final answer, including on the held-out complex-reasoning query, where the correct country and margin appear inside a truncated generation. Perception-only fine-tuning recovers most of these without any reasoning-skill supervision. One probe moves the other way: on the trend question the baseline's free-text response content is correct while the fine-tune leaks, so this sample shows recovery and regression together. Table~\ref{tab:trace-pie-hero} traces every probe.

\begin{table}[htbp]
\centering
\caption{Complex-reasoning accuracy (fuzzy \%) on the held-out CR-500 sets
(bar / pie / CLEVR, 500 questions each), InternVL-3.5-8B. Upper block:
standard skill setups (mean over seeds 42/43/44). Lower block: the two
combined setups retrained at the single-skill budget (2700/300, seed~42).
Baseline is zero-shot (3 eval reps). Subscripts give absolute change over
baseline (\textcolor{impup}{$\uparrow$} gain, \textcolor{impdn}{$\downarrow$}
drop). Best per column per block in \textbf{bold}.}
\label{tab:volume_matched_cr_500}
\setlength{\tabcolsep}{4pt}\small
\begin{tabular}{@{}l ccc@{}}
\toprule
Train setup & Bar & Pie & CLEVR \\
\midrule
Base (no finetune) & 24.0 & 26.0 & 48.0 \\
\emph{P}    & \gainb{36.7}{12.7} & \gainb{40.9}{14.9} & \gain{64.5}{16.5} \\
\emph{G}    & \gain{36.5}{12.5}  & \gain{39.4}{13.4}  & \gainb{67.1}{19.1} \\
\emph{SR}   & \gain{36.7}{12.7}  & \gain{39.5}{13.5}  & \gain{64.5}{16.5} \\
\emph{P+G}  & \gain{28.5}{4.5}   & \gain{34.6}{8.6}   & \gain{62.4}{14.4} \\
\emph{All}  & \gain{35.6}{11.6}  & \gain{40.6}{14.6}  & \gain{62.6}{14.6} \\
\midrule
\rowcolor{cprow}\multicolumn{4}{@{}l}{\textit{Volume-matched (2700/300)}} \\
\emph{P+G}$^{\dagger}$ & \gainb{30.5}{6.5} & \gainb{36.4}{10.4} & \gainb{62.9}{14.9} \\
\emph{All}$^{\dagger}$ & \gain{28.2}{4.2}  & \gain{28.1}{2.1}   & \gain{62.4}{14.4} \\
\bottomrule
\end{tabular}
\end{table}

\begin{table*}[htbp]
\centering
\caption{Per-question diagnostic trace for the bar-chart example, InternVL-3.5-8B baseline vs.\ Perception-only fine-tune. Status: \emph{ok} = both correct; \emph{fixed} = baseline wrong, FT correct; \emph{miss} = wrong under both. A baseline answer that is correct in content but unparseable scores as incorrect, so recovering it counts as \emph{fixed}. \cpgood{Green} marks a scored-correct answer, \cpbad{red} an incorrect one, and \cpfmt{amber} a response whose content is correct but does not commit to a scoreable answer. $^{\dagger}$Q4 (bounding box) is excluded from per-sample scoring; the row is shown for completeness.}
\label{tab:trace-bar-hero}
\begin{minipage}[c]{0.38\textwidth}
\centering
\cpsafeimg{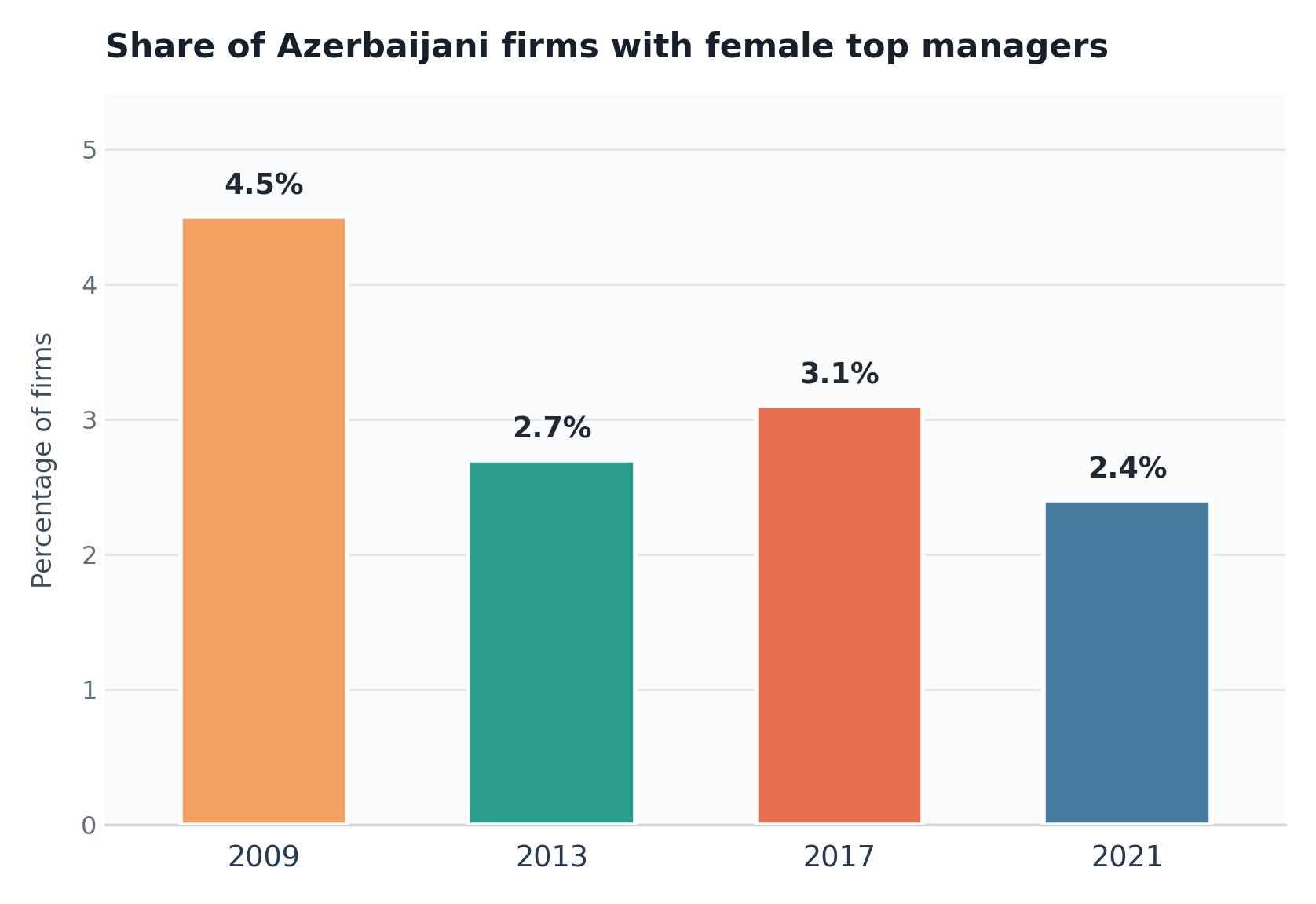}{\linewidth}
\end{minipage}\hfill
\begin{minipage}[c]{0.58\textwidth}
\footnotesize
\textit{Chart overview.} A four-bar horizontal chart of the share of
Azerbaijani firms with female top managers, by survey year: 2009 (4.5\%),
2013 (2.7\%), 2017 (3.1\%), and 2021 (2.4\%). The series fluctuates rather
than trending monotonically; 2009 is the highest and 2021 the lowest, the
mean is $3.175$, and the four bars total $12.7$. Each probe below is
instantiated from this underlying data table.
\end{minipage}
\vspace{6pt}
\setlength{\tabcolsep}{5pt}\footnotesize
\begin{tabularx}{\textwidth}{@{}c >{\raggedright\arraybackslash}X
  >{\raggedright\arraybackslash}p{2.8cm}
  >{\raggedright\arraybackslash}p{2.9cm}
  >{\raggedright\arraybackslash}p{2.9cm} c@{}}
\toprule
\# & Question & GT & Baseline & Perception FT & Status \\
\midrule
\multicolumn{6}{@{}l}{\textit{Perception}}\\
1 & How many bars appear to be above the average height? \newline\cptmpl{P-COUNT-ABOVE-AVG} & \textbf{1} & 1 & 1 & \cppok \\
2 & Is the tallest bar more than twice the height of the shortest bar? \newline\cptmpl{P-HEIGHT-2X} & \textbf{No} & No & No & \cppok \\
3 & What is the overall trend of the bars from top to bottom? \newline\cptmpl{P-TREND-PATTERN} & \textbf{Neither incr.\ nor decr.} & \cpbad{``decrease with fluctuations''} & \cpgood{Neither incr.\ nor decr.} & \cppflip \\
\addlinespace[2pt]
\multicolumn{6}{@{}l}{\textit{Grounding}}\\
4$^{\dagger}$ & What is the bounding box of the bar for 2009? \newline\cptmpl{G-BBOX-ELEMENT} & \textbf{[241, 145, 2197, 381]} & \cpbad{free-text response} & [240, 145, 2100, 350] & \emph{excl.} \\
5 & What is the Year of the topmost bar? \newline\cptmpl{G-VT-LEFTMOST} & \textbf{2009} & 2009 & 2009 & \cppok \\
6 & What is the Percentage of firms for 2017? \newline\cptmpl{G-TV-VALUE-B} & \textbf{3.1} & 3.1\% & 3.1\% & \cppok \\
\addlinespace[2pt]
\multicolumn{6}{@{}l}{\textit{Simple reasoning}}\\
7 & What are the top 3 Years by Percentage of firms? \newline\cptmpl{R-RANK-TOP3} & \textbf{2009, 2017, 2013} & 2009, 2017, 2013 & 2009, 2017, 2013 & \cppok \\
8 & What is the total Percentage of firms across all bars combined? \newline\cptmpl{R-ARITH-TOTAL} & \textbf{12.7} & 12.7\% & 12.7 & \cppok \\
9 & How much higher is 2009 compared to the average Percentage of firms? \newline\cptmpl{R-MULTI-DIFF-MEAN} & \textbf{1.32} \newline {\scriptsize\color{cpmuted}(exact $1.325$)} & \cpbad{1.4\%} & \cpgood{1.3\%} & \cppflip \\
\addlinespace[2pt]
\multicolumn{6}{@{}l}{\textit{Complex reasoning (held-out)}}\\
10 & Which year shows the highest percentage, and by how much does it exceed the lowest year? \newline\cptmpl{COMPLEX} & \textbf{2009; 2.1\%} & \cpbad{free-text response } & \cpgood{2009, 2.1\%} & \cppflip \\
\bottomrule
\end{tabularx}
\end{table*}

\begin{table*}[htbp]
\centering
\caption{Per-question diagnostic trace for the pie-chart example, Qwen3.5-9B baseline vs.\ Perception-only fine-tune. Status: \emph{ok} = both correct; \emph{fixed} = baseline wrong, FT correct; \emph{miss} = wrong under both. A baseline answer that is correct in content but unparseable scores as incorrect, so recovering it counts as \emph{fixed}. \cpgood{Green} marks a scored-correct answer, \cpbad{red} an incorrect one. The Baseline column reports the model's actual concluding statement. Scoring follows the conventions above. $^{\dagger}$Q4 (bounding box) is excluded from per-sample scoring; the row is shown for completeness.}
\label{tab:trace-pie-hero}
\begin{minipage}{\textwidth}
\centering
\begin{minipage}[c]{0.26\textwidth}
  \centering
  \cpsafeimg{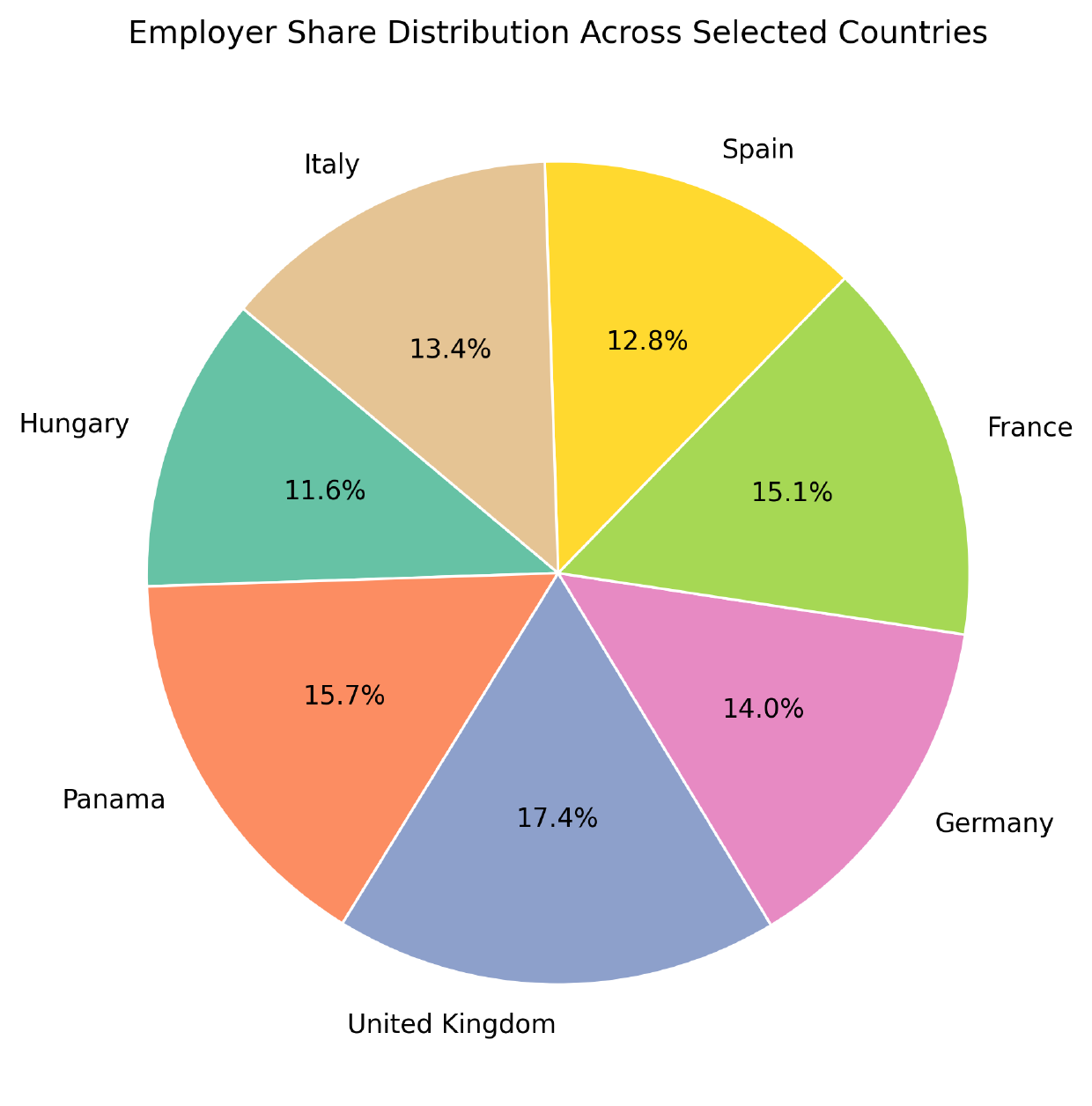}{3.2cm}
\end{minipage}\hfill
\begin{minipage}[c]{0.70\textwidth}
  \footnotesize\textit{Chart overview.} A seven-segment pie of employer
  share across countries: United Kingdom (17.4\%), Panama (15.7\%),
  France (15.1\%), Germany (14.0\%), Italy (13.4\%), Spain (12.8\%), and
  Hungary (11.6\%), summing to 100\%. The largest segment is United
  Kingdom and the smallest is Hungary. Each probe below is instantiated
  from this underlying data table.
\end{minipage}
\end{minipage}
\vspace{4pt}
\setlength{\tabcolsep}{5pt}\footnotesize
\begin{tabularx}{\textwidth}{@{}c 
>{\raggedright\arraybackslash}X
 >{\raggedright\arraybackslash}p{3.2cm}
  >{\raggedright\arraybackslash}p{3cm}
  >{\raggedright\arraybackslash}p{3cm} c@{}}
\toprule
\# & Question & GT & Baseline & Perception FT & Status \\
\midrule
\multicolumn{6}{@{}l}{\textit{Perception}}\\
1 & Segments above average size? \newline\cptmpl{P-COUNT-ABOVE-AVG}
  & \textbf{3} & \cpbad{reasoning, no clean token} & \textcolor{cpok}{3} & \cppflip \\
2 & Largest segment $> 2\times$ smallest? \newline\cptmpl{P-HEIGHT-2X}
  & \textbf{No} & \cpbad{largest ($17.4\%$) less than twice smallest} & \textcolor{cpok}{No} & \cppflip \\
3 & Overall clockwise trend? \newline\cptmpl{P-TREND-PATTERN}
  & \textbf{Neither incr.\ nor decr.} & \textcolor{cpok}{fluctuates, no clear trend} & \cpbad{increasing} & \cppmiss \\
\addlinespace[2pt]
\multicolumn{6}{@{}l}{\textit{Grounding}}\\
4$^{\dagger}$ & Bounding box of UK segment? \newline\cptmpl{G-BBOX-ELEMENT}
  & \textbf{[540, 945, 1335, 1610]} & \cpbad{declines: needs visual coords} & [297,503,684,1000] & \emph{excl.} \\
5 & Country of largest segment? \newline\cptmpl{G-VT-TALLEST}
  & \textbf{United Kingdom} & United Kingdom & United Kingdom & \cppok \\
6 & Percentage for Panama? \newline\cptmpl{G-TV-VALUE-B}
  & \textbf{15.7} & 15.7\% & 15.7\% & \cppok \\
\addlinespace[2pt]
\multicolumn{6}{@{}l}{\textit{Simple reasoning}}\\
7 & Top 3 countries by percentage? \newline\cptmpl{R-RANK-TOP3}
  & \textbf{UK, Panama, France} & \textcolor{cpok}{United Kingdom, Panama, France} & \textcolor{cpok}{UK, Panama, France} & \cppok \\
8 & Total percentage across segments? \newline\cptmpl{R-ARITH-TOTAL}
  & \textbf{100} & 100\% & \textcolor{cpok}{100\%} & \cppok \\
9 & UK vs.\ average percentage? \newline\cptmpl{R-MULTI-DIFF-MEAN}
  & \textbf{3.11} & \cpbad{reasoning, no clean token} & \textcolor{cpok}{3.1} & \cppflip \\
\addlinespace[2pt]
\multicolumn{6}{@{}l}{\textit{Complex reasoning (held-out)}}\\
10 & Highest share and margin over lowest? \newline\cptmpl{COMPLEX}
  & \textbf{UK; 5.8\%} & \cpbad{UK has the highest share at \emph{(cut off)}} & \textcolor{cpok}{United Kingdom, 5.8\%} & \cppflip \\
\bottomrule
\end{tabularx}
\end{table*}

\subsection{CLEVR scene sample}
\label{sec:hero-clevr}
This CLEVR sample shows the effect at its cleanest. Every probe the baseline
answers incorrectly is recovered after perception-only fine-tuning, and no
correct answer regresses. Several of the baseline errors (Q1, Q7, Q8, Q9) are
not perceptual at all: the model already holds the right content but states it
in free-text response rather than committing to a scoreable token. Q2 is the one
substantive perception fix, where the baseline makes an unterminated reasoning
trace and returns no answer. The payoff is the held-out complex-reasoning
question, never trained on, which the fine-tune answers correctly. Table~\ref{tab:trace-clevr} traces the full
sample.

\begin{table*}[htbp]
\centering
\caption{Per-question diagnostic trace for the CLEVR example (\cptid{CLEVR\_val\_014247}), InternVL-3.5-8B baseline vs.\ Perception-only fine-tune. Status: \emph{ok} = both correct; \emph{fixed} = baseline wrong, FT correct; \emph{miss} = wrong under both. A baseline answer that is correct in content but unparseable scores as incorrect, so recovering it counts as \emph{fixed}. \cpgood{Green} marks a scored-correct answer, \cpbad{red} an incorrect one, and \cpfmt{amber} a response whose content is correct but does not commit to a scoreable answer. $^{\dagger}$Q4 (bounding box) is excluded from per-sample scoring; the row is shown for completeness.}
\label{tab:trace-clevr}
\begin{minipage}[c]{0.34\textwidth}
\centering
\cpsafeimg{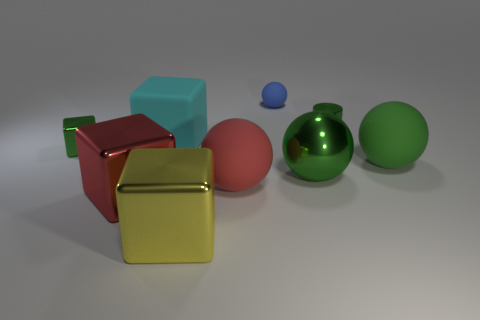}{\linewidth}
\end{minipage}\hfill
\begin{minipage}[c]{0.62\textwidth}
\footnotesize
\textit{Scene overview.} A CLEVR render containing nine objects, six of them
large and four of them spheres. Metal objects outnumber rubber ones. The green
cube and the red cube share the same material, and the red cube lies to the
right of the green cube. Each probe below is instantiated from the scene graph,
so the ground truth is exact rather than estimated from pixels.
\end{minipage}
\setlength{\tabcolsep}{5pt}\footnotesize
\begin{tabularx}{\textwidth}{@{}c >{\raggedright\arraybackslash}X
  >{\raggedright\arraybackslash}p{2.7cm}
  >{\raggedright\arraybackslash}p{3.0cm}
  >{\raggedright\arraybackslash}p{3.0cm} c@{}}
\toprule
\# & Question & GT & Baseline & Perception FT & Status \\
\midrule
\multicolumn{6}{@{}l}{\textit{Perception}}\\
1 & How many objects are in the scene? \newline\cptmpl{P-COUNT-TOTAL} & \textbf{9} & \cpbad{free-text response } & \cpgood{9} & \cppflip \\
2 & How many large objects are in the scene? \newline\cptmpl{P-COUNT-LARGE} & \textbf{6} & \cpbad{unterminated chain, no answer} & \cpgood{6} & \cppflip \\
3 & How many spheres are in the scene? \newline\cptmpl{P-COUNT-SPHERE} & \textbf{4} & \cpgood{4} & \cpgood{4} & \cppok \\
\addlinespace[2pt]
\multicolumn{6}{@{}l}{\textit{Grounding}}\\
4$^{\dagger}$ & What is the bounding box of the green cube? \newline\cptmpl{G-BBOX-ELEMENT} & \textbf{[144, 612, 193, 538]} & \cpbad{free-text response, no box} & {[650, 321, 714, 398]} & \emph{excl.} \\
5 & What color is the green cube? \newline\cptmpl{G-ATTR-COLOR} & \textbf{green} & \cpgood{green} & \cpgood{Green} & \cppok \\
6 & Is the red cube to the left or to the right of the green cube? \newline\cptmpl{G-SPATIAL-LR} & \textbf{right} & \cpgood{right} & \cpgood{right} & \cppok \\
\addlinespace[2pt]
\multicolumn{6}{@{}l}{\textit{Simple reasoning}}\\
7 & Are there more large objects than small objects? \newline\cptmpl{R-COMPARE-LARGE-SMALL} & \textbf{yes} & \cpbad{free-text response } & \cpgood{Yes} & \cppflip \\
8 & Are there more metal objects or rubber objects? \newline\cptmpl{R-MATERIAL-COUNT} & \textbf{metal} & \cpbad{free-text response } & \cpgood{metal} & \cppflip \\
9 & Is the green cube made of the same material as the red cube? \newline\cptmpl{R-SAME-MATERIAL} & \textbf{yes} & \cpbad{free-text response } & \cpgood{Yes} & \cppflip \\
\addlinespace[2pt]
\multicolumn{6}{@{}l}{\textit{Complex reasoning (held-out)}}\\
10 & The metallic thing right of the big shiny object to the right of the big yellow metallic block is what color? \newline\cptmpl{COMPLEX} & \textbf{green} & \cpbad{unterminated chain, no answer} & \cpgood{Green} & \cppflip \\
\bottomrule
\end{tabularx}
\end{table*}

\section{Volume-Matched Ablation}
The combined configurations use more training data than the single-skill ones, so their gains could in principle reflect data volume rather than skill combination. We retrain the two combined setups for InternVL at the single-skill budget (2700/300) and compare (Table~\ref{tab:volume_matched_cr_500}). Perception-plus-grounding retains its advantage under the matched budget, improving slightly on all three domains, so its effect is not a volume artifact. The all-skill configuration, by contrast, loses ground under matching, most visibly on pie, indicating that part of its advantage does come from the larger budget. We therefore read perception-plus-grounding as the more robust combined setup and treat the all-skill results as volume-inclusive.

\end{document}